\documentclass[10pt,twocolumn,letterpaper]{article}

 \usepackage[pagenumbers]{cvpr} 

\usepackage{microtype}

\definecolor{cvprblue}{rgb}{0.21,0.49,0.74}
\usepackage[pagebackref,breaklinks,colorlinks,allcolors=cvprblue]{hyperref}

\usepackage{booktabs}
\usepackage{multirow}
\usepackage{makecell}
\usepackage{colortbl}
\usepackage{xcolor}
\usepackage{algorithm}
\usepackage{algorithmicx}
\usepackage{algpseudocode}
\usepackage{graphicx}

\def\paperID{*****} 
\def\confName{CVPR}
\def\confYear{2026}

\title{Multi-Grained Vision-Language Alignment for\\ Domain Generalized Person Re-Identification}

\author{Jiachen Li$^1$ and Xiaojin Gong\thanks{Corresponding author.}\\
College of Information Science and Electronic Engineering\\
Zhejiang University\\
{\tt\small $^1$12031134@zju.edu.cn, $^*$gongxj@zju.edu.cn}
\and
Dongping Zhang$^2$\\
College of Information Engineering\\
China Jiliang University\\
{\tt\small $^2$06A0303103@cjlu.edu.cn}
}

\begin{document}
\maketitle
\begin{abstract}
	Domain Generalized person Re-identification (DG Re-ID) is a challenging task, where models are trained on source domains but tested on unseen target domains. Although previous pure vision-based models have achieved significant progress, the performance remains further improved. Recently, Vision-Language Models (VLMs) present outstanding generalization capabilities in various visual applications. However, directly adapting a VLM to Re-ID shows limited generalization improvement. This is because the VLM only produces with global features that are insensitive to ID nuances. To tacle this problem, we propose a CLIP-based multi-grained vision-language alignment framework in this work. Specifically, several multi-grained prompts are introduced in language modality to describe different body parts and align with their counterparts in vision modality. To obtain fine-grained visual information, an adaptively masked multi-head self-attention module is employed to precisely extract specific part features. To train the proposed module, an MLLM-based visual grounding expert is employed to automatically generate pseudo labels of body parts for supervision. Extensive experiments conducted on both single- and multi-source generalization protocols demonstrate the superior performance of our approach. The implementation code will be released at \url{https://github.com/RikoLi/MUVA}.
\end{abstract}

\section{Introduction}

Person Re-identification (Re-ID)~\cite{BoT,TransReID,Tran-GCN} has been extensively studied to match identities across multiple non-overlapped cameras. In despite of considerable advances, conventional Re-ID models degenerate when tested in new scenarios due to domain gaps. Consequently, Domain Generalized Person Re-identification (DG Re-ID) has been proposed, which learns models on source domains and is tested on unseen target domains.

Over few past years, various pure vision-based paradigms have been proposed to tackle DG Re-ID, including feature disentanglement and normalization~\cite{SNR,DIR-ReID,CBN}, mixture-of-experts~\cite{META,RaMoE,SALDG}, meta-learning~\cite{MetaBIN,M3L,MDA,SuA-SpML} and so on. Although significant progress has been achieved, DG Re-ID remains challenging to apply in real-world scenarios. This is because pure visual information contains limited generalization knowledge. In an image, domain-specific information (\eg illumination, view angle, image resolution and so on) tends to tightly entangle with ID-specific information. When models are trained, the ID recognition learning is accompanied by the learning of domain-specific information. Such models may perform well on test samples from source domains, but they degenerate drastically when transferred to quite different scenarios due to significant gaps of vision data distribution.

\begin{figure*}[tp]
	\centering
	\includegraphics[width=0.8\textwidth]{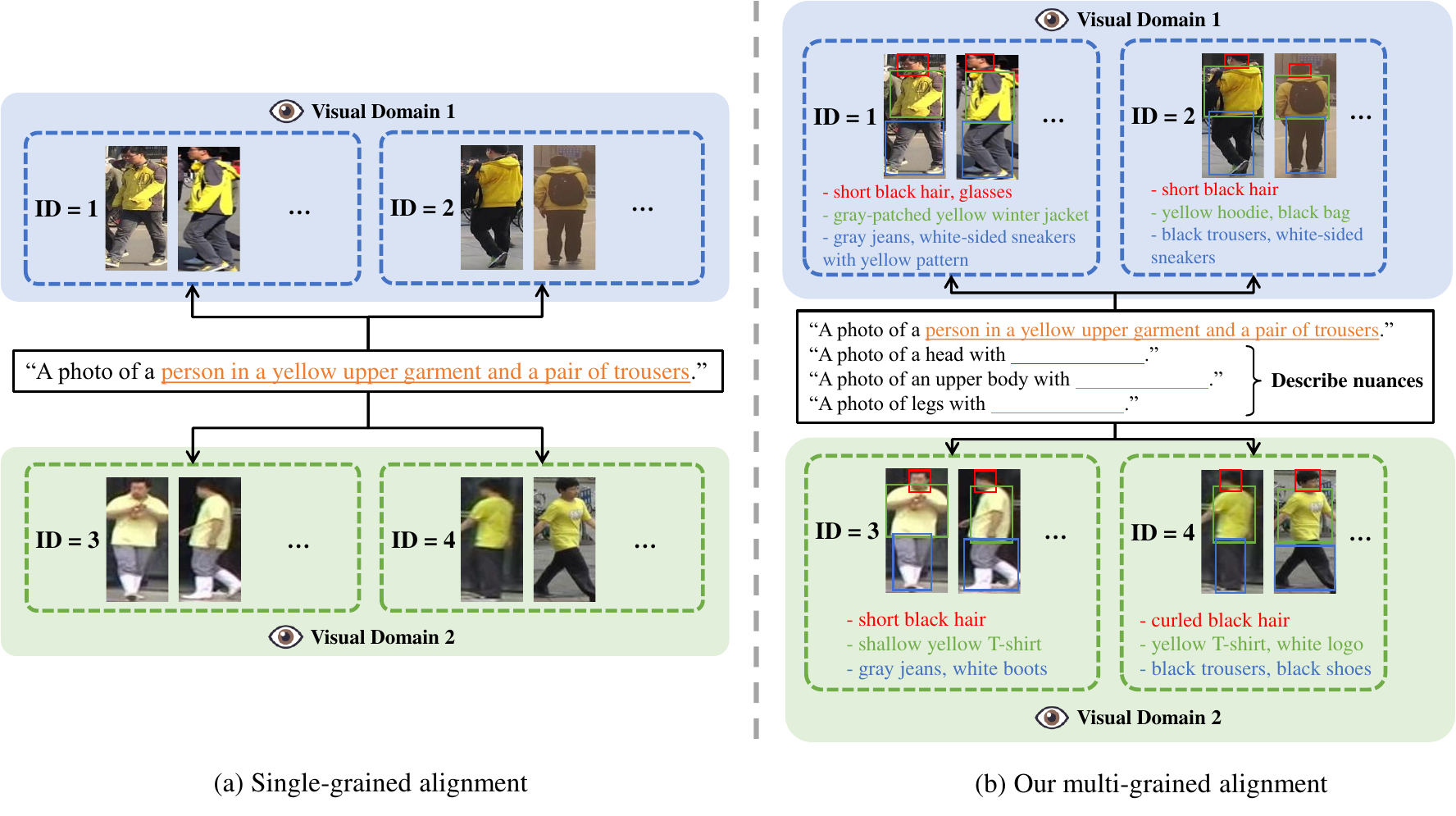}
	\caption{Comparison of the single- and multi-grained alignments. (a) indicates the single-grained alignment. Similar IDs from diverse visual domains can be summarized by a sentence in a global view. Domain-specific attributes such as illumination, view angle, image resolution and so on are eliminated due to the nature of language. However, the single-grained description ignores ID nuances. (b) indicates our multi-grained alignment, where extra fine-grained descriptions of parts are employed to represent more subtle ID differences. Note that exact words in descriptions are used as examples to explain our idea. In implementation, several learnable prompts are adopted for more flexible descriptions instead of manually designed words.}
	\label{fig:idea}
\end{figure*}

Recently, Vision-Language Models (VLMs)~\cite{CLIP,ALIGN} perform remarkably well on out-of-distribution data due to the utilization of language modality in vision tasks. Inspired by this, we expect that the potential of DG Re-ID can be further excavated if the VLM is adopted to align vision and language modalities. As illustrated in Figure~\ref{fig:idea} (a), each ID can be summarized by a textual description. The description is inherently ID-relevant. Domain-specific attributes are naturally eliminated when each image is aligned with text. But some methods~\cite{DMRL,MMET,CLIP-FGDI} still obtain suboptimal performances even if they have included text in training. This is because they merely exploit single-grained information. Although text is generalizable, it is simultaneously challenged to represent the nuances among similar IDs. Details like hairstyles, clothes patterns and so on are not sufficiently represented in single-grained manner. However, these detailed cues are crucial for precise recognition, which have been extensively utilized in many previous Re-ID tasks~\cite{PCB,MGN,TMGF,AAformer,PASS,ge2024semantically,yu2025no}. Therefore, a multi-grained alignment between two modalities is expected as illustrated in Figure~\ref{fig:idea} (b), which additionally focuses on local details of body parts and describes the ID nuances comprehensively.

To complete multi-grained alignment, it is important to extract local features at different granularities. Most previous approaches adopt stripe-like partition~\cite{PCB,MGN,TMGF} to equally divide an image into stripes along height direction. This strategy is simple and easy to implement. However, the perception field of each rigid stripe is constrained to a fixed size, making it unable to flexibly capture local features on deformable body parts, especially when a person is walking or varying body poses. This tends to damage the vision-language alignment of nuanced semantics. A better approach to adaptively extract local features is required to mitigate this problem.

We propose a MUlti-grained Vision-language Alignment (MUVA) framework by adapting a well-known VLM model CLIP~\cite{CLIP} to solve aforementioned issues within a unified architecture. To address the nuance perception deficiency, we design a multi-grained vision-language prompt learning strategy. Specifically, we propose learnable multi-grained prompts, which are particularly designed to align both the entire human body and its local parts. To sufficiently adapt CLIP~\cite{CLIP}, a two-stage training pipeline is adopted. Initially, the learnable prompts of each ID are trained to align with corresponding images, obtaining multi-grained textual descriptions. After that, the descriptions offer a cross-modal constraint during Re-ID training to learn generalizable features.

To align with textual descriptions, local visual features at multiple granularities are extracted. Instead of using rigid stripes, we propose an Adaptively Masked Multi-head Self-Attention (AM-MSA) module to improve CLIP~\cite{CLIP}'s image encoder for precise part feature extraction. In each module, a transformer is equipped with a bypass branch containing a Residual Mask Prediction (RMP) module and a Mask Gating (MG) module. The RMP module predicts specific part masks layer by layer with cross-attention mechanism. The MG module further employs masked attention on the transformer with the predicted masks to focus on detailed semantics. To train the AM-MSA module, a Visual Grounding Expert (VGE) based on a Multi-modal Large Language Model (MLLM)~\cite{LLaVA,liu2024improved} is used to automatically generate pseudo labels for each part, thereby avoiding tedious manual annotation. In inference, the AM-MSA module can adaptively predict body parts, enabling our model to accurately capture both global context and fine-grained local features. In summary, our major contributions are as follows:
\begin{itemize}
	\item We adapt vision-language alignment to improve DG Re-ID with the inherent generalization capability of language. To cope with the nuance perception deficiency in single-grained alignment, we propose a multi-grained vision-language alignment framework based on a multi-grained prompt learning strategy.
	\item To precisely extract local information, we design an adaptively masked multi-head self-attention module to predict part regions. To train this module without tedious manual annotation, an MLLM-based visual grounding expert is employed to generate pseudo labels for different parts as supervision.
	\item Extensive experiments conducted under both single- and multi-source DG Re-ID protocols demonstrate the effectiveness of our approach, surpassing state-of-the-art methods by a remarkable margin.
\end{itemize}

\section{Related Works}

\subsection{Domain Generalized and Part-Based Person Re-ID}

Domain Generalized (DG) Re-ID approaches can be roughly divided into following categories: feature disentanglement and normalization~\cite{SNR,DIR-ReID,CBN}, learning by mixture-of-experts~\cite{META,RaMoE,SALDG}, meta-learning~\cite{MetaBIN,M3L,MDA,SuA-SpML}, image matching~\cite{QAConv,QAConv-GS,QAConv-MS,TransMatcher} and unsupervised pre-training~\cite{ISR}. Besides, some new methods like MLFP~\cite{MLFP} and U-DG~\cite{U-DG} adopt perturbation-based data augmentation on input images or intermediate features, encouraging robust feature learning. These approaches mainly focus on pure visual information. Recently, the language modality has been gradually introduced to improve generalizability. MMET~\cite{MMET} and DMRL~\cite{DMRL} introduce large-scale pre-training on a synthesized multi-modal dataset FineGPR~\cite{FineGPR}. CLIP-FGDI~\cite{CLIP-FGDI} learns generalizable features with domain-invariant/relevant prompts based on CLIP~\cite{CLIP}. However, they are challenged to distinguish subtle ID differences since only global features are utilized.

Part-based approaches are widely adopted to reduce ID ambiguity with fine-grained information. In traditional intra-domain Re-ID, multi-grained structure design~\cite{MGN,zheng2019pyramidal}, attention-based methods~\cite{Li2018HAN,Si2018Dual,AiT} and transformer-based methods~\cite{AAformer,PASS,TMGF,GAIN} have been proven effective. Initially, stripe-like partition~\cite{PCB,Cheng2016} is adopted extensively to extract fine-grained features from fixed regions. But they suffer from precisely locating body parts due to varying body poses. Afterwards, PASS~\cite{PASS} introduces random cropping to capture diverse local views based on fixed regions. AAformer~\cite{AAformer} regards the part discovery as an optimal transport problem between part tokens and image patches. These approaches have been demonstrated effective but the improvements are still limited. Differently, our approach introduces an extra module to adaptively locate parts, which is supervised by pseudo labels generated by a visual grounding expert. The expert offers sufficient prior knowledge of human body to train the proposed module, endowing the module with more flexible and accurate locating capability in inference. Assisted by the proposed module, nuanced visual information in parts can be perceived and aligned with language modality in a multi-grained manner, which significantly enhances the performance of DG Re-ID.

\subsection{CLIP and Prompt Learning}
Contrastive Language-Image Pre-training (CLIP)~\cite{CLIP} is a seminal work in vision-language alignment. Leveraging the knowledge learned from a large-scale dataset collected from the Internet, it exhibits outstanding generalization capacity. However, re-training CLIP~\cite{CLIP} from scratch on specific downstream tasks is impractical due to the enormous computational resources required.

Prompt learning has emerged as an efficient approach to adapt CLIP to downstream tasks, which freezes the model encoders and only optimizes some learnable prompts, including both single-modal~\cite{CoOp,CoCoOp,KgCoOp} and multi-modal~\cite{MaPLe} settings. Notably, CLIP-ReID~\cite{CLIP-ReID} firstly adapts prompt learning to intra-domain supervised Re-ID. But it aligns two modalities only in a global view, neglecting the fine-grained information that is important to ID distinguishment. In this paper, we notice the deficiency of global alignment, thus introducing a multi-grained alignment pipeline in both global and local views, and exploring the performance of this pipeline in DG Re-ID.

\subsection{MLLM-based Visual Grounding}
Traditional object detection approaches~\cite{FasterRCNN,YOLO,DETR} predict bounding boxes regressively. Different from them, visual grounding aims to locate specific image regions in language given textual descriptions. Currently, it has witnessed significant progress driven by the advancement of Multi-modal Large Language Models (MLLMs)~\cite{LLaVA,liu2024improved}. As a representative work, Shikra~\cite{Shikra} directly outputs object coordinates in natural language. Ferret series~\cite{Ferret,Ferret-v2} extend the grounding ability to arbitrary granularities, shapes and resolutions. LLaVA-Grounding~\cite{LLaVA-Grounding} adopts diverse input and output forms. In this work, we adopt a visual grounding model as an expert to annotate part locations for images before training. After that, these locations serve as pseudo labels to supervise a proposed module for precise part locating in training. We select CogVLM~\cite{CogVLM} as the expert considering its superior grounding performance among multiple rivals~\cite{Shikra,Ferret,Qwen-VL}.

\begin{figure*}[t]
	\centering
	\includegraphics[width=\textwidth]{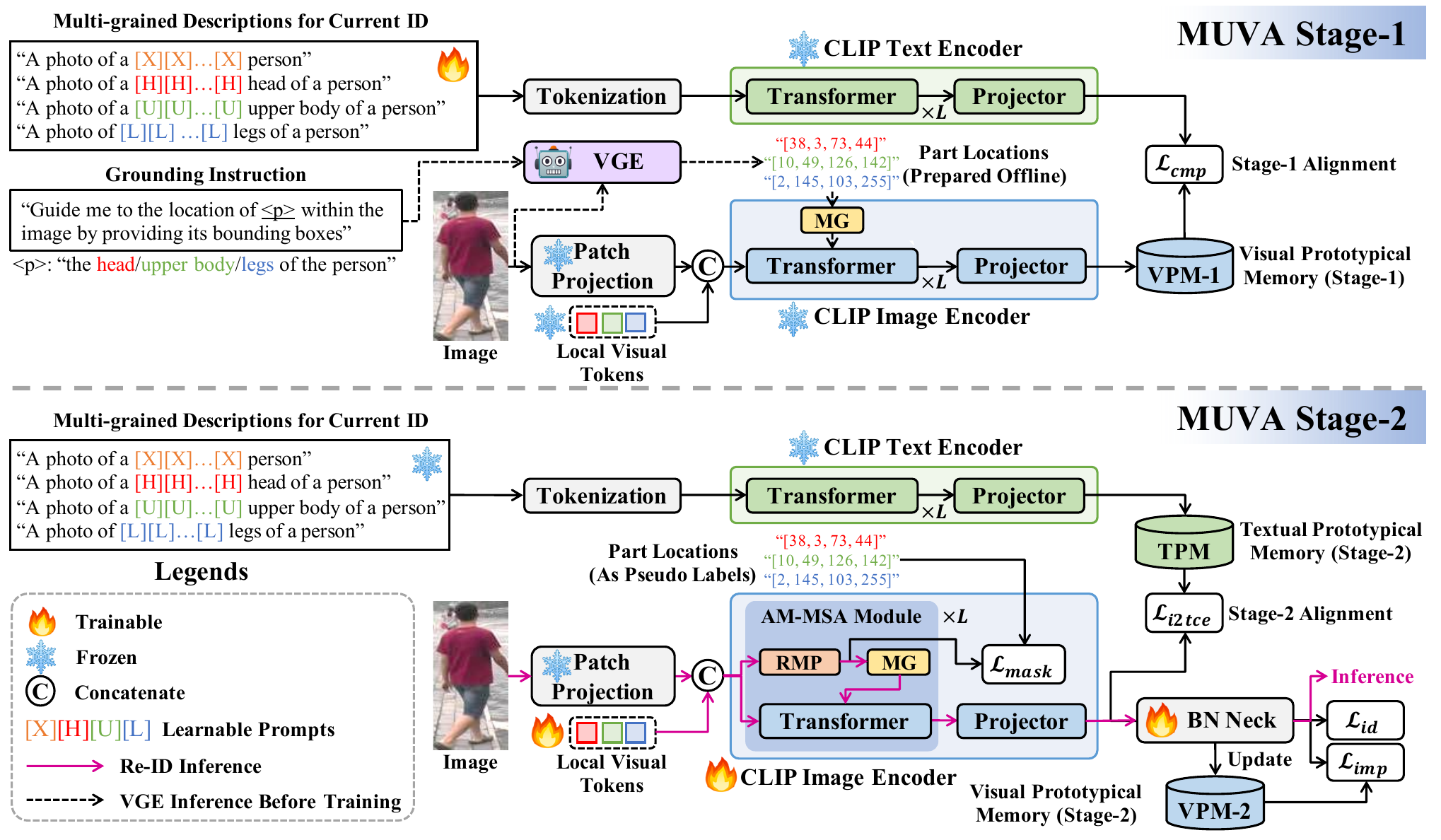}
	\caption{Pipeline of our framework. Before training, the VGE generates part locations for upcoming usage. In stage-1, the multi-grained learnable prompts of each ID are aligned with correponding images. The part locations are directly used to extract local visual features. In stage-2, the proposed AM-MSA module is adopted in each transformer layer to extract local features and optimized with the entire image encoder. In this procedure, the part locations act as pseudo labels to supervise the AM-MSA module. Prototypical memories are utilized to train two stages. In inference, only the image encoder is utilized.
	}
	\label{fig:pipeline}
\end{figure*}

\subsection{Masked Attention Mechanism}
Considering the development of transformer architecture~\cite{Transformer}, attention mechanism~\cite{OrientedFormer,RQFormer} has been adopted broadly. Among various attention designs, masked attention shows its practicability in multiple fields. In natural language processing, the stair-like causal mask~\cite{Transformer,GPT3} is adopted to predict next token with previous contents, ensuring the causality of generated contents. In computer vision, Mask2Former~\cite{Mask2Former} introduces masked attention to focus on specific instances. CamoFormer~\cite{CamoFormer} directly applies pixel-level masks on multi-scale feature maps before attention computation to better detect camouflaged objects. In recommendation algorithms, PMAN~\cite{PMAN} and AMA-CDR~\cite{AMA-CDR} utilize attention masks to erase noisy information passed in graph models. In this work, we adopt masked attention to focus on particular body parts, aiming to align them with their counterparts in language modality for generalizable feature learning. The masked attention also helps improving fine-grained feature extraction against varying body poses to offer stronger ID discriminability in our approach.

\section{The Proposed Method}
As illustrated in Figure~\ref{fig:pipeline}, our model builds upon CLIP~\cite{CLIP} image and text encoders. We introduce local visual tokens and learnable multi-grained prompts for finer vision-language alignment via prompt learning. To focus on varying body parts, we introduce an Adaptively Masked Multi-head Self-Attention (AM-MSA) module to locate parts and aggregate fine-grained information precisely. To train the AM-MSA module, a Visual Grounding Expert (VGE) is employed to produce pseudo labels for part locating supervision. The model is optimized in a two-stage training strategy and infers with the image encoder only.

\subsection{Multi-grained Vision-Language Prompt Learning}
\label{sec:MPL}
\subsubsection{Multi-grained Prompts}
Formally, we design multi-grained description templates for each ID in source domain to offer adequate textual contexts. From a global perspective, we inherit the practice in CLIP-ReID~\cite{CLIP-ReID} by adopting the template ``A photo of a $[\text{X}]_1^c \cdots [\text{X}]_N^c$ person''.  Here, $[\text{X}]_n^c \in \mathbb{R}^{D_{prompt}}$ represents a learnable global prompt embedded in the fixed contexts, where $n \in \{1, \cdots, N\}$ is the index of each prompt, $N$ is the number of prompts, $c \in \{1, \cdots, C\}$ is the ID label, $C$ is the total number of IDs and $D_{prompt}$ is prompt dimension. From a local perspective, additional prompts are introduced as ``A photo of a $[\text{H}]_1^c \cdots [\text{H}]_N^c$ head of a person'', ``A photo of a $[\text{U}]_1^c \cdots [\text{U}]_N^c$ upper body of a person'' and ``A photo of $[\text{L}]_1^c \cdots [\text{L}]_N^c$ legs of a person'', corresponding to three major parts of a human body. $[\text{H}]_n^c$, $[\text{U}]_n^c$ and $[\text{L}]_n^c$ refer to learnable local prompts, which maintain the same dimension $D_{prompt}$ with the global prompt. Collectively, the global and local prompts are termed multi-grained prompts. Instead of using exact words to describe each ID, the learnable multi-grained prompts can provide a more informative textual description for each ID in language embedding space. We denote the descriptions of all granularities together as $\mathcal{T}$. After tokenization, the descriptions $\mathcal{T}$ are fed into the text encoder $\mathcal{E}_T$ to obtain a multi-grained textual token $\mathbf{t}^{mg} = \mathcal{E}_T (\mathcal{T})$, where $\mathbf{t}^{mg} \in \mathbb{R}^{d}$ and $d$ denotes its dimension.

For the vision modality, a similar scheme is adopted by introducing additional local visual tokens to fuse fine-grained semantics from corresponding parts. Given the token sequence after image patchification, three newly added local visual tokens are concatenated to the sequence as enhancements for original global visual token. The entire sequence is denoted as $\mathcal{V}$ and then sent to the image encoder $\mathcal{E}_I$ for feature extraction. During this process, the local visual tokens are exploited for fine-grained information aggregation through the AM-MSA module (to be introduced in Section~\ref{sec:AM-MSA}). Finally, a multi-grained visual token is obtained by $\mathbf{v}^{mg} = \mathcal{E}_I (\mathcal{V})$, where $\mathbf{v}^{mg} \in \mathbb{R}^{d}$.

\subsubsection{Two-stage Training}
Inspired by CLIP-ReID~\cite{CLIP-ReID}, our multi-grained vision-language prompt learning framework incorporates two training stages. In stage-1, the learnable prompts of each ID are optimized to align with the correponding images but all other parameters remain frozen. In stage-2, only the visual tokens, the image encoder and a BN neck~\cite{BoT} are optimized for Re-ID feature learning. Simultaneously, the learned prompts are employed to build a cross-modal constraint to improve generalizability using the information of language modality.

Conventionally, instance-level contrastive learning is essential for excellent vision-language alignment~\cite{CLIP,CoOp,CoCoOp,CLIP-ReID}. However, the number of instances is restricted by batch size, leading to insufficient contrasting and suboptimal performance, especially when the computation resource is limited. Therefore, we adopt Prototypical Contrastive Losses (PCL)~\cite{PCL} to conduct prototype-level contrasting, where the number of positive and negative contrastive samples is decoupled with the batch size by utilizing memories~\cite{ClusterContrast,CAP,O2CAP}.

Formally, in stage-1 we use the textual token of each ID to contrast with a Visual Prototypical Memory (VPM-1) $\mathcal{M}^V_1 \in \mathbb{R}^{C \times d}$ for vision-language alignment. For the $c$-th ID, a cross-modal multi-grained prototypical contrastive loss $\mathcal{L}_{cmp}$ is defined as:
\begin{equation}
	\label{eqn:cmp_loss}
	\mathcal{L}_{cmp} = - \log \frac{\exp\left(\mathrm{sim}\left(\mathbf{t}^{mg}_c, \mathcal{M}^V_1[c]\right)\right)}{\sum_{j=1}^C \exp\left(\mathrm{sim}\left(\mathbf{t}^{mg}_c, \mathcal{M}^V_1[j]\right)\right)},
\end{equation}
where $\mathbf{t}^{mg}_c = \mathcal{E}_T(\mathcal{T}_c)$ denotes the normalized multi-grained textual token after encoding the $c$-th ID's description $\mathcal{T}_c$. $\mathrm{sim}(\cdot, \cdot)$ denotes cosine similarity. Each entity $\mathcal{M}^V_1[c]$ in the VPM-1 is a normalized centroid of all visual tokens labeled as $c$, which is computed in advance and prepared to be directly pulled close towards its corresponding textual token $\mathbf{t}^{mg}_c$. During this process, the ID information in the image is squeezed out in the form of language, which is more generalizable, and preserved in the learnable prompts by gradient backpropagation through the text encoder.

In stage-2, the Re-ID learning is conducted to extract discriminative features of different IDs. We adopt ID loss $\mathcal{L}_{id}$ as usual but omit the triplet loss~\cite{tripletloss} used in most previous approaches for contrastive learning. Instead, we adopt an intra-modal multi-grained prototypical contrastive loss $\mathcal{L}_{imp}$ for enhanced contrasting, assisted by another Visual Prototypical Memory (VPM-2) $\mathcal{M}^V_2 \in \mathbb{R}^{C \times d}$ containing the visual token centroids of all IDs. Given the $i$-th sample:
\begin{equation}
	\begin{aligned}
		\label{eqn:reid_loss}
		\mathcal{L}_{id} &= - \log \frac{\exp\left(\mathbf{w}_{y_i}^\top\mathbf{v}_i^{mg}\right)}{\sum_{j=1}^C \exp\left(\mathbf{w}_j^\top\mathbf{v}_i^{mg}\right)}, \\
		\mathcal{L}_{imp} &= - \log \frac{\exp\left(\mathrm{sim}\left(\mathbf{v}_i^{mg}, \mathcal{M}^V_2[y_i]\right) / \tau \right)}{\sum_{j=1}^C \exp\left(\mathrm{sim}\left(\mathbf{v}_i^{mg}, \mathcal{M}^V_2[j]\right) / \tau \right)},
	\end{aligned}
\end{equation}
where $\mathbf{w}_{j} \in \mathbb{R}^d$ denotes the classifier for the $j$-th ID, $\tau$ denotes the temperature regulating contrastive strength, and $\mathbf{v}_i^{mg}$ and $y_i$ denote the visual token and the ID label of the $i$-th sample, respectively. Unlike the VPM-1, VPM-2 is updated with a momentum factor $\gamma$ to keep pace with the rapid update of the image encoder:
\begin{equation}
	\label{eqn:memory_update}
	\mathcal{M}^V_2[y_i] \leftarrow \gamma \mathcal{M}^V_2[y_i] + (1 - \gamma) \mathbf{v}^{mg*},
\end{equation}
where $\mathbf{v}^{mg*}$ is the hardest sample~\cite{ClusterContrast} for each ID in the batch. Compared with the triplet loss~\cite{tripletloss}, the memory-based $\mathcal{L}_{imp}$ significantly enlarges the number of contrastable samples to all IDs, thereby achieving better performance. Meanwhile, the image-to-text cross-entropy loss $\mathcal{L}_{i2tce}$ is adopted to introduce generalizable knowledge from language modality:
\begin{equation}
	\label{eqn:i2tce_loss}
	\mathcal{L}_{i2tce} = - \sum_{c=1}^C q_c \log \frac{\exp\left(\mathrm{sim}\left(\mathbf{v}_i^{mg}, \mathcal{M}^T[c]\right)\right)}{\sum_{j=1}^C \exp\left(\mathrm{sim}\left(\mathbf{v}_i^{mg}, \mathcal{M}^T[j]\right)\right)},
\end{equation}
where $q_c$ denotes the smoothed ID label. $\mathcal{M}^T \in \mathbb{R}^{C \times d}$ is a Textual Prototypical Memory (TPM) initialized with normalized textual token $\mathbf{t}^{mg}_c$ for each ID label $c$. It is built by encoding the prompts learned in stage-1. On one hand, the Re-ID learning using $\mathcal{L}_{id}$ and $\mathcal{L}_{imp}$ ensures the model's discrimination capability. On the other hand, $\mathcal{L}_{i2tce}$ enables the injection of the language knowledge squeezed in stage-1 into the image encoder, thereby enhancing the model's generalization capability.

\subsection{Adaptively Masked Multi-Head Self-Attention Module}
\label{sec:AM-MSA}
To obtain fine-grained part features, most traditional strategies~\cite{PCB,MGN,TMGF,LA-Transformer,PAT} split the image into fixed stripes. It is straightforward but fails to keep the semantic alignment between body parts and the stripe-like regions, especially when the body is in varying poses. To address this issue, we propose an Adaptively Masked Multi-head Self-Attention (AM-MSA) module based on the typical transformer structure. This module is capable of precisely discovering parts with foreground mask prediction using cross-attention mechanism.
\begin{figure}[t]
	\centering
	\includegraphics[width=\linewidth]{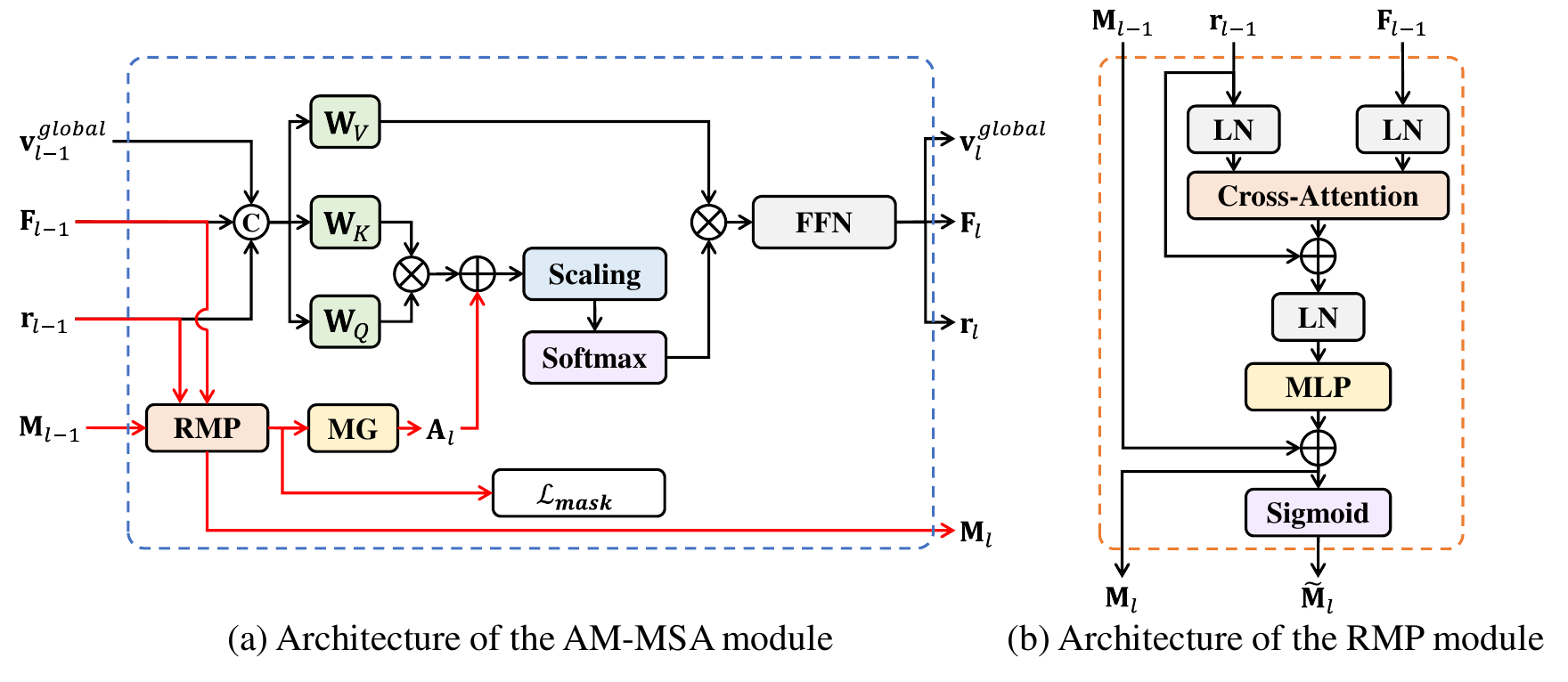}
	\caption{Illustration of the AM-MSA module. (a) indicates its inner architecture. The red path highlights a bypass branch which adaptively generates an attention mask $\mathbf{A}_l$ for local information aggregation in a transformer. For simplicity, the residual connections in the transformer are omitted. (b) indicates the architecture of the RMP module, where a cross-attention layer, an MLP layer, several layer normalization layers and a sigmoid activation function are used to predict the foreground mask of each part.}
	\label{fig:am-msa}
\end{figure}

As illustrated in Figure~\ref{fig:am-msa}, a Residual Mask Prediction (RMP) module is embedded in a bypass branch paralleled to the multi-head self-attention in transformer. Formally, the inputs of the $l$-th layer are the outputs of the $(l-1)$-th layer, including residual part mask scores $\mathbf{M}_{l-1} \in \mathbb{R}^{3 \times N_{patch}}$ for three parts, image patch tokens $\mathbf{F}_{l-1} \in \mathbb{R}^{N_{patch} \times D}$, combined local visual tokens $\mathbf{r}_{l-1} \in \mathbb{R}^{3 \times D}$ and a global visual token $\mathbf{v}_{l-1}^{global} \in \mathbb{R}^{D}$. Here, $N_{patch}$ and $D$ denote the number of image patches and the feature dimension in the image encoder, respectively. $\mathbf{r}_{l-1}$ is obtained by concatenating all local visual tokens. $\mathbf{v}^{global}_{l-1}$ is directly fetched from the token sequence, which is also know as the ``CLS'' token in vision transformers~\cite{ViT,CLIP}. The outputs of the $l$-th layer are denoted similarly to the inputs but with the subscripts changed to $l$. In each layer, a Mask Gating (MG) module is used to convert the output of the RMP module into an attention mask $\mathbf{A}_l \in \mathbb{R}^{3 \times N_{patch}}$. After that, $\mathbf{A}_l$ is employed in the masked attention mechanism to focus on parts. To train such process, a mask loss is optimized by the pseudo labels produced by an MLLM-based visual grounding expert.

\subsubsection{Residual Mask Prediction Module}
The Residual Mask Prediction (RMP) module is designed to predict foreground masks for different parts. Given the combined local visual tokens $\mathbf{r}_{l-1}$ and the image patch tokens $\mathbf{F}_{l-1}$, the mask $\tilde{\mathbf{M}}_l$ is predicted as follows:
\begin{equation}
	\begin{aligned}
		&\mathbf{r}'_{l-1} = \mathrm{CrossAttn}(\mathrm{LN}(\mathbf{r}_{l-1}), \mathrm{LN}(\mathbf{F}_{l-1})) + \mathbf{r}_{l-1}, \\
		&\mathbf{M}_l = \mathrm{MLP}(\mathrm{LN}(\mathbf{r}'_{l-1})) + \mathbf{M}_{l-1}, \\
		&\tilde{\mathbf{M}}_l = \sigma(\mathbf{M}_l),
	\end{aligned}
\end{equation}
where $\mathrm{CrossAttn}(\cdot, \cdot)$, $\mathrm{LN}(\cdot)$, $\mathrm{MLP}(\cdot)$ and $\sigma(\cdot)$ denote the cross-attention, layer normalization, multi-layer perceptron and sigmoid activation, respectively.
By investigating the correlation between the combined local visual tokens $\mathbf{r}_{l-1}$ and the image patch tokens $\mathbf{F}_{l-1}$, where $\mathbf{r}_{l-1}$ produces the query and $\mathbf{F}_{l-1}$ produces the key and value features, the output $\mathbf{r}'_{l-1} \in \mathbb{R}^{3 \times D}$ collects adequate information to locate parts. The information is then used to predict part mask scores $\mathbf{M}_l \in \mathbb{R}^{3 \times N_{patch}}$ through the MLP. To fuse semantics different layers, we employ residual connections around the RMP module. Part mask scores predicted in previous layer are added to those in next layer, aiming to offer prior information to assist mask prediction in deeper RMP modules. Inside the RMP module, we also add a residual connection from the input to the output of the cross-attention, further enhancing the features used to predict part masks. At the end, a sigmoid activation $\sigma(\cdot)$ is applied to convert the scores into probabilities $\tilde{\mathbf{M}}_l$.

\begin{figure*}[t]
	\centering
	\includegraphics[width=\textwidth]{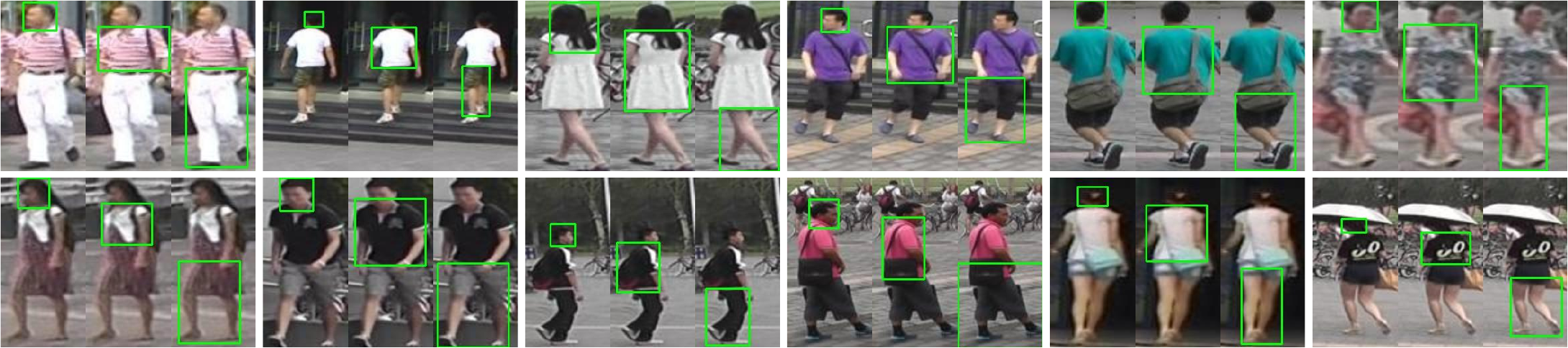}
	\caption{Visualization of part locating with VGE on Market1501~\cite{Market1501} dataset. Different parts are detected in green boxes.}
	\label{fig:vis_vge_results}
\end{figure*}

\subsubsection{Mask Gating Module}
The Mask Gating (MG) module converts the foreground masks predicted by the RMP into attention masks. Then, the masked attention mechanism is adopted to extract detailed information from specific regions. This is completed by filling a new matrix $\mathbf{A}_l$ as follows:
\begin{equation}
	\label{eqn:mask_gating}
	A_{i,l}^p = \begin{cases}
		0 & \text{if } \tilde{M}_{i,l}^p > t\\
		-\infty & \text{otherwise} \\
	\end{cases},
\end{equation}
where $A_{i,l}^p$ is a value in $\mathbf{A}_l$, representing the attention mask of the $i$-th image patch belonging to part $p$, predicted by the $l$-th layer. $\tilde{M}_{i,l}^p$ is a value in $\tilde{\mathbf{M}}_l$, sharing the same super- and subsrcipts with $A_{i,l}^p$. $t$ denotes a decision threshold. $\mathbf{A}_l$ is added to the original Multi-head Self-Attention (MSA) scores in the transformer after being padded to the same size:
\begin{equation}
	\mathrm{MSA}(\mathbf{Q}_l, \mathbf{K}_l, \mathbf{V}_l) = \mathrm{softmax}\left( \frac{\mathbf{Q}_l \mathbf{K}_l^\top}{\sqrt{D}} + \mathrm{Pad}(\mathbf{A}_l) \right) \mathbf{V}_l,
\end{equation}
where $\mathbf{Q}_l$, $\mathbf{K}_l$ and $\mathbf{V}_l$ are query, key and value matrices of the $l$-th layer, repectively. The padding operation is denoted as $\mathrm{Pad}(\cdot)$. The softmax function will eliminate irrelated tokens with score $-\infty$, allowing the local visual tokens to focus on specific parts for fine-grained information aggregation.

\begin{figure}[t]
	\centering
	\includegraphics[width=\linewidth]{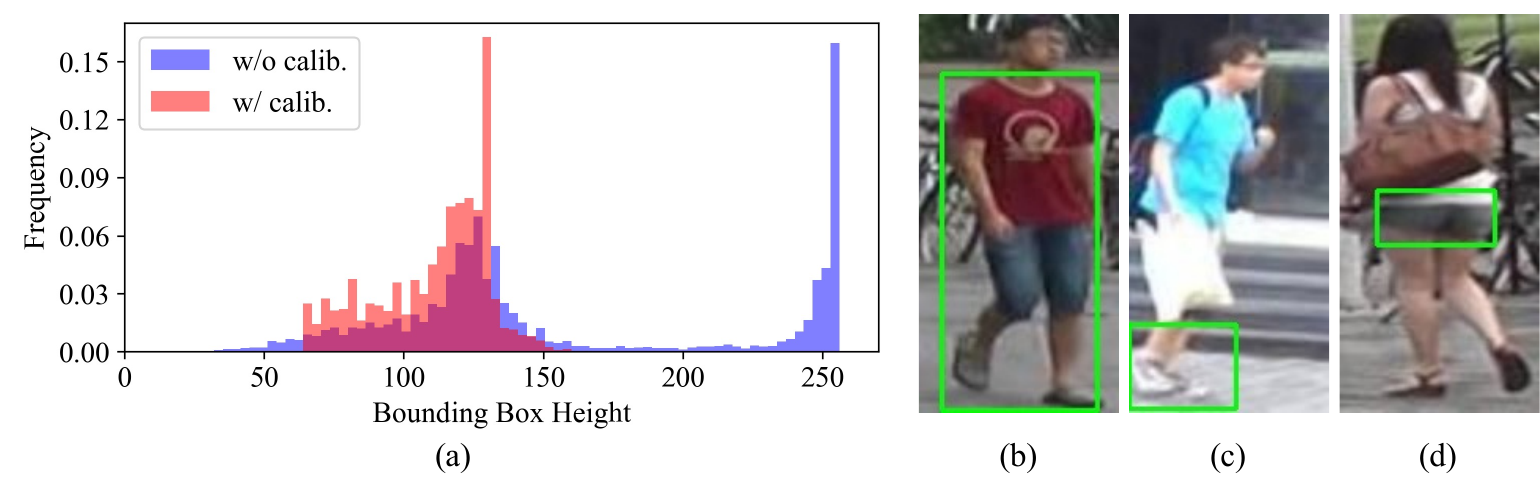}
	\caption{Illustration of the imperfect results. In (a), the distribution of the height of bounding boxes for leg regions from Market1501~\cite{Market1501} dataset is presented. Blue denotes the raw results and red denotes the calibrated results. Without calibration, we find irregular peaks at values approximating the entire image height, which correspond to incorrect locations containing the whole person body. Extremely small values are also incorrect, which often represent over-splitting. (b) is a sample of oversized partition. (c) and (d) are examples of over-split leg regions. Green boxes are predictions of the VGE.}
	\label{fig:market1501_leg_bbox_height_calib_compare}
\end{figure}

\begin{figure*}[t]
	\centering
	\includegraphics[width=\textwidth]{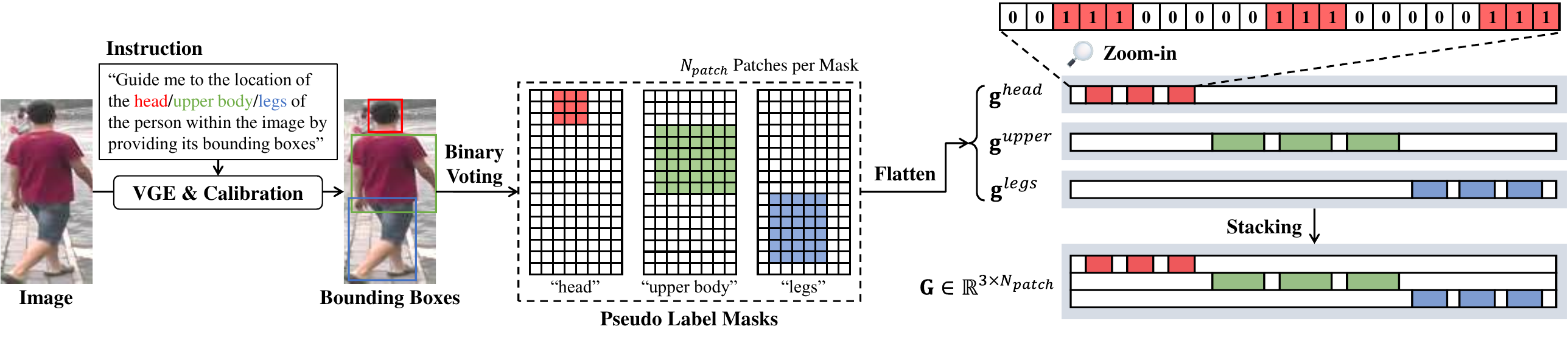}
	\caption{Illustration of the generation of pseudo label masks. The masks are generated by a binary voting strategy. After that, the masks are flattened in row-wise and stacked together.}
	\label{fig:ground_truth_local_mask_generation}
\end{figure*}

\subsubsection{Visual Grounding Expert Supervision}
\label{sec:VGE}
To train the AM-MSA module, we employ an MLLM-based Visual Grounding Expert (VGE) as a supervisor. Although the VGE locates parts precisely, it is computationally expensive when applied on-the-fly during inference. Therefore, we utilize the VGE's predictions as pseudo labels to teach the lightweight AM-MSA module to explore parts, enabling acceptable computational cost during inference.

The VGE is constructed upon a recent visual grounding model CogVLM~\cite{CogVLM}, which shows extensive effectiveness over 10 cross-modal benchmarks~\cite{CogVLM}. By virtue of its strong understanding of multi-modal contents, it achieves state-of-the-art performances on typical visual grounding benchmarks~\cite{xiao2025towards} containing RefCOCO/+/g~\cite{nagaraja2016modeling,yu2016modeling} and Visual7W~\cite{zhu2016visual7w} datasets. Additionally, CogVLM~\cite{CogVLM} takes inputs according to a pre-defined prompt template and outputs grounding coordinates directly in language, enabling convenient inference in an end-to-end manner.

Formally, a language instruction $\mathcal{Q}^p$ ``Guide me to the location of $<$p$>$ within the image by providing its bounding boxes'' is utilized as a query to prompt the expert $\mathcal{E}_{VGE}$ with a person image $\mathbf{I}$. The placeholder ``$<$p$>$'' is replaced with descriptions of different body parts, i.e. ``the head/upper body/legs of the person''. Given such inputs, the VGE generates an answer containing the bounding box coordinates $\mathbf{b} = [x_{min}, y_{min}, x_{max}, y_{max}]$ for each part. This process is formally described as $\mathbf{b}^p = \mathcal{E}_{VGE}(\mathbf{I}, \mathcal{Q}^p)$, where the superscript $p \in \{ head, upper, legs \}$ denotes each possible part name. Consequently, we obtain the part locations $\mathbf{b}^{head}$, $\mathbf{b}^{upper}$ and $\mathbf{b}^{legs}$ corresponding to the instruction queries $\mathcal{Q}^{head}$, $\mathcal{Q}^{upper}$ and $\mathcal{Q}^{legs}$, respectively. Some grounding results are visualized in Figure~\ref{fig:vis_vge_results}.

However, there still remains some challenging cases where the parts are inaccurately located due to the hallucination of large language models~\cite{HALoGEN,huang2023survey}, as illustrated in Figure~\ref{fig:market1501_leg_bbox_height_calib_compare}. To address this problem, we calibrate inferior predictions based on statistical charateristics of bounding boxes. Specifically, extremely large or small boxes are rectified with stripe-like windows that focus on the top, middle or bottom areas of the entire image since corresponding parts are likely to appear in these regions. The calibration is employed independently for each part as $\tilde{\mathbf{b}}^{p} = \mathrm{Calib} (\mathbf{b}^{p})$, where $\mathrm{Calib}(\cdot)$ denotes the calibration operation and $\tilde{\mathbf{b}}^p$ represents the calibrated bounding box.

Subsequently, we generate pseudo labels with the calibrated bounding boxes. Given a rectified bounding box $\tilde{\mathbf{b}}^p$, a binary voting strategy is employed to generate a pseudo label mask of each part by checking the positions of image patch tokens as illustrated in Figure~\ref{fig:ground_truth_local_mask_generation}. If a patch is covered by the bounding box, its position is filled by ``1'' in the mask. Otherwise, ``0'' is filled. Each mask is flattened in row-wise
and denoted as $\mathbf{g}^p \in \mathbb{R}^{N_{patch}}$ for part $p$. Finally, the masks of all parts are stacked together to form a pseudo label mask matrix $\mathbf{G} \in \mathbb{R}^{3 \times N_{patch}}$.

The pseudo label masks are used to supervise the part discovery in the AM-MSA module by a combination of binary cross-entropy loss and dice loss~\cite{dice_loss}. For each sample:
\begin{equation}
	\begin{aligned}
		&\mathcal{L}_{bce} = \frac{1}{3N_{patch}L} \sum_{p, i, l} \mathrm{BinaryCrossEntropy}(\tilde{M}_{i,l}^p, G_i^p), \\
		&\mathcal{L}_{dice} = \frac{1}{3N_{patch}L} \sum_{p, i, l} \left( 1 - \frac{2  \tilde{M}_{i, l}^p  G_i^p}{\tilde{M}_{i, l}^p + G_i^p} \right),
	\end{aligned}
\end{equation}
where $\tilde{M}_{i,l}^p \in [0, 1]$ (a value of $\tilde{\mathbf{M}}_l$ predicted in the AM-MSA module) represents the probability of the $i$-th image patch belonging to part $p$, predicted by the $l$-th layer. $G_i^p \in \{ 0, 1 \}$ is the corresponding value of $\mathbf{G}$ as ground truth. $L$ denotes the total number of layers in the model. The losses are averaged over each region, image patch and layer. The binary cross-entropy loss $\mathcal{L}_{bce}$ is designed to locate the correct parts while the dice loss~\cite{dice_loss} is introduced to enhance the consistency between the predicted masks and the pseudo label masks. The overall mask loss $\mathcal{L}_{mask}$ is defined as:
\begin{equation}
	\label{eqn:mask_loss}
	\mathcal{L}_{mask} = \mathcal{L}_{bce} + \mathcal{L}_{dice}.
\end{equation}

Optimized by the mask loss, the AM-MSA module learns to locate body parts by itself, making it available to automatically discover and extract fine-grained information during inference without the online use of VGE.

\begin{algorithm}[tb]
	\caption{MUVA Training.}
	\label{alg:muva}
	\renewcommand{\algorithmicrequire}{\textbf{Inputs:}}
	\renewcommand{\algorithmicensure}{\textbf{Outputs:}}
	
	\begin{algorithmic}[1]
		\Require Image encoder $\mathcal{E}_{I}$, text encoder $\mathcal{E}_{T}$, visual grounding expert $\mathcal{E}_{VGE}$, vision token sequence $\mathcal{V}$ after image patchification, multi-grained text descriptions $\mathcal{T}$ (including learnable prompts) after tokenization, number of IDs $C$, training epochs for stage-1 $S_1$ and training epochs for stage-2 $S_2$
		\Ensure Trained image encoder $\mathcal{E}_I$ and BN neck
		
		\State Prepare pseudo label mask matrices $\mathbf{G}$ for training samples via $\mathcal{E}_{VGE}$
		\State Set learnable prompts in $\mathcal{T}$ as trainable
		\State Create a memory $\mathcal{M}_1^V$
		\For {$epoch$ in $[1, \cdots, S_1]$} \Comment{Stage-1 training}
		\For{$c$ in $[1,\cdots, C]$}
		\State Compute $\mathbf{t}^{mg}_c = \mathcal{E}_T(\mathcal{T}_c)$
		\State Compute $\mathcal{L}_{cmp}$ in Equation~\eqref{eqn:stage1_loss} with $\mathbf{t}^{mg}_c$ and $\mathcal{M}^V_1$
		\State Backpropagate and update learnable prompts in $\mathcal{T}_c$
		\EndFor
		\EndFor
		\State Create a memory $\mathcal{M}^T$ with updated $\mathcal{T}$ in stage-1
		\State Set $\mathcal{E}_I$ and BN neck as trainable
		\For{$epoch$ in $[1, \cdots, S_2]$} \Comment{Stage-2 training}
		\State Create a memory $\mathcal{M}_2^V$
		\For{$batch$}
		\State Compute $\mathbf{v}^{mg} = \mathcal{E}_I(\mathcal{V})$
		\State Compute $\mathcal{L}_{id}$, $\mathcal{L}_{imp}$, $\mathcal{L}_{i2tce}$ and $\mathcal{L}_{mask}$ in Equation~\eqref{eqn:stage2_loss} with $\mathbf{v}^{mg}$, $\mathcal{M}^V_2$, $\mathcal{M}^T$ and $\mathbf{G}$
		\State Backpropagate and update parameters of $\mathcal{E}_I$ and BN neck
		\State Update $\mathcal{M}^V_2$ with Equation~\eqref{eqn:memory_update}
		\EndFor
		\EndFor
		
		\State \Return $\mathcal{E}_I$ and BN neck
	\end{algorithmic}
\end{algorithm}

\subsection{The Entire Training Loss}
In general, our approach is summarized in Algorithm~\ref{alg:muva}. As preprocessing, the VGE predicts the pseudo label masks of all samples in source domain training set in advance, preparing for upcoming utilization. In stage-1, $\mathcal{L}_{cmp}$ is optimized to train the learnable prompts. The training loss of this stage is formulated as:
\begin{equation}
	\label{eqn:stage1_loss}
	\mathcal{L}_{stage1} = \mathcal{L}_{cmp}.
\end{equation}

In stage-2, the Re-ID learning is carried out by optimizing $\mathcal{L}_{id}$ and $\mathcal{L}_{imp}$, assisted by the image-to-text cross-entropy loss $\mathcal{L}_{i2tce}$ for generalizable knowledge injection from the prompts learned before. To provide with precise fine-grained information during training, the visual tokens building the memory $\mathcal{M}^V_2$ are also obtained with the help of VGE-predicted masks. Meanwhile, the AM-MSA module is supervised by adopting the mask loss $\mathcal{L}_{mask}$ to learn part locating. The training losses of this stage are formulated as:
\begin{equation}
	\label{eqn:stage2_loss}
	\mathcal{L}_{stage2} = \mathcal{L}_{id} + \mathcal{L}_{imp} + \mathcal{L}_{i2tce} + \mathcal{L}_{mask}.
\end{equation}

\section{Experiments}
\subsection{Datasets and Evaluation Protocols}

Our experiments are conducted on following large-scale datasets: Market1501~\cite{Market1501}, DukeMTMC-reID~\cite{DukeMTMC-reID}, MSMT17~\cite{MSMT17} and CUHK03-NP~\cite{re-ranking}, abbreviated as MA, D, MS and C3, respectively. As usual, the DG Re-ID evaluation is performed under both single-source and multi-source protocols. For the single-source generalization, the model is trained only on training set of one dataset and evaluated on testing set of the other dataset. This is more challengeable since the number of source domain is restricted to one. For the multi-source generalization, we take a leave-one-out strategy on MA, D, MS and C3 like previous approaches, where the testing set of one dataset is selected to be evaluated, while the mixture of the training sets of other datasets are used to train the model. For all evaluation protocols, the Mean Average Precision (mAP) and Cumulative Matching Characteristic (CMC) at Rank-1 are reported as numerical metrics for Re-ID performance. No re-ranking~\cite{re-ranking} techniques are used in evaluation.

\begin{table*}[tp]
	\centering
	\caption{Comparison with state-of-the-art performances under single-source protocol. $*$ indicates that the input image size is larger than 256$\times$128. $\dagger$ indicates that the method uses target domain data for test-time updating. $\ddagger$ indicates that the method uses extra datasets for pre-training. The results marked in \textcolor{gray}{gray} are produced by running the source code provided by authors, and all others are directly adopted from their original papers. Best and suboptimal results are emphasized in bold and underline, respectively.}
	\label{tab:ssdg_sota}
	\resizebox{\textwidth}{!}{
		\begin{tabular}{lccccccccccccc}
			\toprule
			\multirow{2}{*}{Model} & \multirow{2}{*}{Type} & \multicolumn{2}{c}{MA$\to$D} & \multicolumn{2}{c}{MA$\to$MS} & \multicolumn{2}{c}{D$\to$MA} & \multicolumn{2}{c}{D$\to$MS} & \multicolumn{2}{c}{MS$\to$MA} & \multicolumn{2}{c}{MS$\to$D} \\ \cmidrule{3-14} 
			&  & mAP & Rank-1 & mAP & Rank-1 & mAP & Rank-1 & mAP & Rank-1 & mAP & Rank-1 & mAP & Rank-1 \\ \midrule
			QAConv$^*$~\cite{QAConv}$_\textit{ECCV'20}$ & \multirow{4}{*}{Image Matching} & 28.7 & 48.8 & 7.0 & 22.6 & 27.2 & 58.6 & 8.9 & 29.0 & 43.1 & 72.6 & 52.6 & 69.4 \\
			TransMatcher$^*$~\cite{TransMatcher}$_\textit{NeurIPS'21}$ & & \textcolor{gray}{42.6} & \textcolor{gray}{61.8} & 18.4 & 47.3 & \textcolor{gray}{46.9} & \underline{\textcolor{gray}{77.5}} & \textcolor{gray}{20.7} & \textcolor{gray}{52.8} & 52.0 & 80.1 & \textcolor{gray}{50.8} & \textcolor{gray}{67.9} \\
			QAConv-GS$^*$~\cite{QAConv-GS}$_\textit{CVPR'22}$ & & \textcolor{gray}{39.3} & \textcolor{gray}{60.5} & 17.2 & 45.9 & \textcolor{gray}{41.1} & \textcolor{gray}{72.6} & \textcolor{gray}{17.0} & \textcolor{gray}{46.9} & 49.5 & 79.1 & \textcolor{gray}{47.5} & \textcolor{gray}{67.3} \\
			QAConv-MS$^*$~\cite{QAConv-MS}$_\textit{TCSVT'24}$ & & - & - & 19.9 & 49.7 & - & - & - & - & 54.5 & \bf 83.0 & - & - \\
			\midrule
			PAT~\cite{PAT}$_\textit{ICCV'23}$ & Part-based & 48.9 & 67.9 & 18.2 & 42.8 & 45.2 & 71.9 & 19.2 & 43.9 & 47.3 & 72.2 & \textcolor{gray}{54.9} & \textcolor{gray}{71.4} \\ \midrule
			MetaBIN~\cite{MetaBIN}$_\textit{CVPR'21}$ & \multirow{3}{*}{Meta-Learning} & 33.1 & 55.2 & \textcolor{gray}{10.4} & \textcolor{gray}{31.0} & 35.9 & 69.2 & \textcolor{gray}{12.1} & \textcolor{gray}{37.5} & \textcolor{gray}{39.9} & \textcolor{gray}{70.4} & \textcolor{gray}{42.2} & \textcolor{gray}{64.2} \\
			MDA$\dagger$~\cite{MDA}$_\textit{CVPR'22}$ & & 34.4 & 56.7 & 11.8 & 33.5 & 38.0 & 70.3 & - & - & 53.0 & 79.7 & \textcolor{gray}{52.4} & \textcolor{gray}{71.7} \\
			SuA-SpML~\cite{SuA-SpML}$_\textit{TIP'23}$ &  & 34.8 & 55.5 & 11.1 & 30.1 & 36.3 & 65.8 & 13.6 & 37.8 & - & - & - & - \\ \midrule
			CBN~\cite{CBN}$_\textit{ECCV'20}$ & \multirow{9}{*}{Normalization} & 38.2 & 58.7 & 9.5 & 25.3 & 43.0 & 72.7 & 13.0 & 35.4 & 45.0 & 73.7 & 46.7 & 66.2 \\
			DTIN-Net~\cite{DTIN-Net}$_\textit{ECCV'22}$ &  & 36.1 & 57.0 & - & - & 37.4 & 69.8 & - & - & - & - & - & - \\
			GN~\cite{GN}$_\textit{TCSVT'23}$ &  & 34.0 & 52.3 & 10.3 & 28.6 & 34.3 & 64.3 & 12.3 & 33.8 & - & - & - & - \\
			GN+SNR~\cite{GN}$_\textit{TCSVT'23}$ &  & 34.7 & 55.4 & - & - & 36.9 & 68.5 & - & - & 37.5 & 68.0 & 45.4 & 66.2 \\
			STL~\cite{STL}$_\textit{ICME'24}$ &  & - & - & 19.8 & 48.9 & - & - & - & - & 54.0 & 82.5 & - & - \\
			LDU~\cite{LDU}$_\textit{TIM'24}$ &  & 38.0 & 59.5 & 13.5 & 35.7 & 42.3 & 73.2 & 16.7 & 44.2 & 44.8 & 74.6 & 48.9 & 69.2 \\
			MTI~\cite{MTI}$_\textit{JVCIR'24}$ &  & 36.4 & 57.8 & - & - & 38.2 & 70.5 & - & - & 42.7 & 72.9 & 47.7 & 67.5 \\
			SVIL~\cite{SVIL}$_\textit{TOMM'24}$ &  & 35.1 & 56.0 & - & - & 35.1 & 67.9 & - & - & - & - & - & - \\
			IGMG$^*$~\cite{IGMG}$_\textit{CVIU'24}$ &  & 41.6 & 64.2 & - & - & 42.1 & 75.8 & - & - & 53.5 & 80.9 & \underline{61.6} & \underline{77.7} \\ \midrule
			MLFP~\cite{MLFP}$_\textit{TIM'25}$ & Augmentation & 38.8 & 60.8 & 11.9 & 34.3 & 36.9 & 69.7 & 12.1 & 35.3 & - & - & - & - \\ \midrule 
			DMRL$\ddagger$~\cite{DMRL}$_\textit{Mach. Learn.'24}$ & \multirow{5}{*}{VLM-based} & - & - & 21.5 & 50.6 & - & - & - & - & \underline{55.1} & 81.3 & - & - \\
			MMET$\ddagger$~\cite{MMET}$_\textit{TOMM'25}$ &  & - & - & \underline{26.8} & 50.1 & - & - & - & - & 40.6 & 76.3 & - & - \\
			CLIP-FGDI$^*$~\cite{CLIP-FGDI}$_\textit{TIFS'25}$ &  & \textcolor{gray}{41.6} & \textcolor{gray}{64.1} & \textcolor{gray}{20.2} & \textcolor{gray}{47.0} & \underline{\textcolor{gray}{48.6}} & \textcolor{gray}{75.7} & \underline{\textcolor{gray}{25.2}} & \underline{\textcolor{gray}{53.8}} & \textcolor{gray}{44.7} & \textcolor{gray}{73.0} & \textcolor{gray}{47.5} & \textcolor{gray}{66.9} \\
			DCAC~\cite{DCAC}$_\textit{Sensors'25}$ &  & \underline{49.5} & \underline{69.1} & 23.4 & \underline{52.1} & 42.3 & 71.5 & 19.7 & 47.4 & 52.1 & 77.9 & 58.4 & 75.0 \\
			MUVA$_\textit{Ours}$ &  & \bf 52.0 & \bf 72.1 & \bf 27.6 & \bf 56.8 & \bf 51.4 & \bf 78.3 & \bf 26.8 & \bf 55.7 & \bf 59.5 & \underline{82.7} & \bf 62.0 & \bf 77.8 \\ \bottomrule
	\end{tabular}}
\end{table*}

\begin{table*}[tp]
	\centering
	\caption{Additional performances under single-source protocol on CUHK03-NP~\cite{re-ranking}. $*$ indicates that the input image size is larger than 256$\times$128. $\ddagger$ indicates that the method uses extra datasets for pre-training. The results marked in \textcolor{gray}{gray} are produced by running the source code provided by authors, and all others are directly adopted from their original papers. Best and suboptimal results are emphasized in bold and underline, respectively.}
	\label{tab:ssdg_c3_sota}
	\resizebox{\textwidth}{!}{
		\begin{tabular}{lccccccccccccc}
			\toprule
			\multirow{2}{*}{Model} & \multirow{2}{*}{Type} & \multicolumn{2}{c}{MA$\to$C3} & \multicolumn{2}{c}{D$\to$C3} & \multicolumn{2}{c}{MS$\to$C3} & \multicolumn{2}{c}{C3$\to$MA} & \multicolumn{2}{c}{C3$\to$D} & \multicolumn{2}{c}{C3$\to$MS} \\ \cmidrule{3-14} 
			&  & mAP & Rank-1 & mAP & Rank-1 & mAP & Rank-1 & mAP & Rank-1 & mAP & Rank-1 & mAP & Rank-1 \\ \midrule
			QAConv~\cite{QAConv}$_\textit{ECCV'20}$ & \multirow{4}{*}{Image Matching} & 8.6 & 9.9 & 6.8 & 7.9 & 22.6 & 25.3 & \textcolor{gray}{20.1} & \textcolor{gray}{45.9} & \textcolor{gray}{15.0} & \textcolor{gray}{29.4} & \textcolor{gray}{3.9} & \textcolor{gray}{14.9} \\
			TransMatcher$^*$~\cite{TransMatcher}$_\textit{NeurIPS'21}$ &  & 21.4 & 22.2 & \textcolor{gray}{18.2} & \textcolor{gray}{18.9} & 22.5 & 23.7 & \textcolor{gray}{38.8} & \textcolor{gray}{69.7} & \textcolor{gray}{34.6} & \textcolor{gray}{56.8} & \textcolor{gray}{15.2} & \textcolor{gray}{45.2} \\
			QAConv-GS$^*$~\cite{QAConv-GS}$_\textit{CVPR'22}$ &  & 18.1 & 19.1 & \textcolor{gray}{14.7} & \textcolor{gray}{15.1} & 20.6 & 20.9 & \textcolor{gray}{37.9} & \textcolor{gray}{69.0} & \textcolor{gray}{34.9} & \textcolor{gray}{57.5} & \textcolor{gray}{16.2} & \textcolor{gray}{47.6} \\
			QAConv-MS$^*$~\cite{QAConv-MS}$_\textit{TCSVT'24}$ &  & 24.9 & 26.6 & - & - & 28.5 & 31.0 & - & - & - & - & - & - \\ \midrule
			PAT~\cite{PAT}$_\textit{ICCV'23}$ & Part-based & 26.0 & 25.4 & 18.9 & 18.8 & - & - & - & - & - & - & - & - \\ \midrule
			GN~\cite{GN}$_\textit{TCSVT'23}$ & \multirow{5}{*}{Normalization} & 14.5 & 14.4 & 10.3 & 10.2 & - & - & 40.6 & 67.6 & 31.2 & 50.0 & 11.9 & 33.4 \\
			GN+SNR~\cite{GN}$_\textit{TCSVT'23}$ &  & 15.2 & 15.1 & 11.5 & 11.0 & 18.3 & 17.4 & - & - & - & - & - & - \\
			STL~\cite{STL}$_\textit{ICME'24}$ &  & 26.9 & 27.7 & - & - & 24.5 & 25.6 & - & - & - & - & - & - \\
			LDU~\cite{LDU}$_\textit{TIM'24}$ &  & 18.2 & 18.5 & 14.2 & 14.2 & 21.3 & 21.3 & 37.5 & 68.1 & 29.5 & 51.8 & 12.6 & 36.9 \\
			MTI~\cite{MTI}$_\textit{JVCIR'24}$ &  & 16.2 & 16.3 & 13.3 & 13.3 & 16.0 & 15.4 & - & - & - & - & - & - \\ \midrule
			MLFP~\cite{MLFP}$_\textit{TIM'25}$ & Augmentation & - & - & - & - & - & - & 39.0 & \bf 70.4 & 33.8 & 57.0 & - & - \\ \midrule
			DMRL$\ddagger$~\cite{DMRL}$_\textit{Mach. Learn.'24}$ & \multirow{3}{*}{VLM-based} & 22.6 & 23.4 & - & - & 24.7 & 26.1 & - & - & - & - & - & - \\
			DCAC~\cite{DCAC}$_\textit{Sensors'25}$ &  & \underline{32.5} & \underline{33.2} & \underline{23.0} & \underline{23.5} & \underline{34.1} & \underline{34.4} & \underline{42.0} & 68.6 & \underline{43.2} & \underline{64.8} & \underline{17.8} & \underline{47.3} \\
			MUVA$_\textit{Ours}$ &  & \bf 38.8 & \bf 40.8 & \bf 29.4 & \bf 31.1 & \bf 40.2 & \bf 42.0 & \bf 45.1 & \underline{70.2} & \bf 45.6 & \bf 66.7 & \bf 21.0 & \bf 49.8 \\ \bottomrule
	\end{tabular}}
\end{table*}

\subsection{Implementation Details}
The model's backbone is built upon CLIP~\cite{CLIP}. To avoid training instability, we freeze the patch projection layer of the image encoder during both two training stages. The input image is resized to 256$\times$128 after random horizontal flipping. The number of learnable prompts $N$ is set to 4 with dimension $D_{prompt} = 512$ to describe each part. The newly introduced local visual tokens share the same positional embedding with the global visual token. The learnable prompts and all newly added visual tokens are randomly initialized from the normal distribution with standard deviations 0.002 and 0.02, respectively. We adopt CogVLM~\cite{CogVLM} with weight \texttt{cogvlm-grounding-generalist-v1.1} as the visual grounding expert, keeping its hyper-parameters unchanged. The threshold $t$ in the MG module is set to 0.5. The number of the cross-attention heads in the RMP modules are adjusted according to the scale of training sets, which is set to 2 for all single-source protocols except for training on MS, and 8 for training on MS and all multi-source protocols.

In stage-1 training, the multi-grained learnable prompts are optimized with an initial learning rate of $3.5\times 10^{-4}$, regulated by a cosine scheduler for 120 epochs. In stage-2 training, the image encoder equipped with the AM-MSA module is optimized with an initial learning rate of $5 \times 10^{-6}$ for 60 epochs, regulated by a step scheduler using linear warmup. The new parameters in the RMP module are optimized with the learning rate multiplied by a scale factor of 10. The visual prototypical memory VPM-2 is updated with a momentum $\gamma = 0.2$. The temperature $\tau$ for $\mathcal{L}_{imp}$ is set to 0.01. In both stages, the Adam~\cite{Adam} optimizer is employed with a weight decay of $1\times 10^{-4}$. The batch size is set to 64 by randomly selecting 16 IDs and 4 samples per ID. The entire approach is implemented via PyTorch~\cite{pytorch}. All of the experiments are conducted on a single NVIDIA RTX A6000 GPU.

\subsection{Comparison with State-of-the-art Methods}

\subsubsection{Single-source DG Re-ID}

Table~\ref{tab:ssdg_sota} and \ref{tab:ssdg_c3_sota} present comprehensive comparisons of our proposed approach under single-source protocols against multiple state-of-the-art methods. The results clearly demonstrate that our MUVA model significantly outperforms most prior approaches. Specifically, our model exhibits superior performance compared to the methods utilizing larger input sizes, such as QAConv series~\cite{QAConv,QAConv-GS,QAConv-MS}, TransMatcher~\cite{TransMatcher} and IGMG~\cite{IGMG}. It also surpasses MDA~\cite{MDA}, which applies test-time updating on target domains. On the source domain MS, recent methods such as QAConv-MS~\cite{QAConv-MS}, STL~\cite{STL} and IGMG~\cite{IGMG} show competitive performances. However, these methods offer limited improvements when trained on other source domains, where our approach continues to exhibit significant enhancements. For Vision-Language Model (VLM)-based approaches, although DMRL~\cite{DMRL} and MMET~\cite{MMET} employ extra datasets for multi-modal pre-training, our MUVA model achieves higher generalization capability using only the source domain data. A recent work CLIP-FGDI~\cite{CLIP-FGDI} extending CLIP-based prompt learning with domain-invariant/relevant prompts achieves a competitive performance on D$\to$MS. But it is challenged to obtain balanced improvements on other protocols. Compared with a diffusion-based model DCAC~\cite{DCAC}, our model also performs significant advantages over multiple protocols without generative learning objectives.

\begin{table*}[tp]
	\centering
	\caption{Comparison with state-of-the-art performances under multi-source protocol. $*$ indicates that the input image size is larger than 256$\times$128. $\dagger$ indicates that the method uses target domain data for test-time updating. $\ddagger$ indicates that the method uses extra datasets for pre-training. The results marked in \textcolor{gray}{gray} are produced by running the source code provided by authors, and all others are directly adopted from their original papers. Best and suboptimal results are emphasized in bold and underline, respectively.}
	\label{tab:msdg_sota}
	\resizebox{\textwidth}{!}{
		\begin{tabular}{lccccccccc}
			\toprule
			\multirow{2}{*}{Model} & \multirow{2}{*}{Type} & \multicolumn{2}{c}{D+MS+C3$\to$MA} & \multicolumn{2}{c}{MA+MS+C3$\to$D} & \multicolumn{2}{c}{MA+D+C3$\to$MS} & \multicolumn{2}{c}{MA+D+MS$\to$C3} \\ \cmidrule{3-10} 
			&  & mAP & Rank-1 & mAP & Rank-1 & mAP & Rank-1 & mAP & Rank-1 \\ \midrule
			QAConv-MS$^*$~\cite{QAConv-MS}$_\textit{TCSVT'24}$ & Image Matching & 65.2 & \bf 87.1 & - & - & 25.9 & \underline{57.7} & 40.3 & 44.0 \\ \midrule
			PAT~\cite{PAT}$_\textit{ICCV'23}$ & Part-based & 51.7 & 75.2 & 56.5 & 71.8 & 21.6 & 45.6 & 31.5 & 31.1 \\ \midrule
			ISR$^*\ddagger$~\cite{ISR}$_\textit{ICCV'23}$ & Pre-training & \bf 70.5 & \underline{87.0} & - & - & \underline{30.3} & 56.4 & 37.8 & 36.6 \\ \midrule
			M$^3$L~\cite{M3L}$_\textit{CVPR'21}$ & \multirow{2}{*}{Meta-Learning} & 48.1 & 74.5 & 50.5 & 69.4 & 12.9 & 33.0 & 29.9 & 30.7 \\
			SuA-SpML~\cite{SuA-SpML}$_\textit{TIP'23}$ &  & 59.1 & 83.2 & 51.9 & 71.6 & 20.4 & 47.1 & 33.3 & 34.4 \\ \midrule
			RaMoE~\cite{RaMoE}$_\textit{CVPR'21}$ & \multirow{2}{*}{Mixture-of-Experts} & 56.5 & 82.0 & 56.9 & 73.6 & 13.5 & 34.1 & 35.5 & 36.6 \\
			SALDG~\cite{SALDG}$_\textit{IJMLC'24}$ &  & 57.6 & 82.3 & 52.0 & 71.2 & 18.1 & 46.5 & 32.4 & 34.5 \\ \midrule
			DEX~\cite{DEX}$_\textit{BMVC'21}$ & \multirow{2}{*}{\makecell{Embedding\\ Expansion}} & 55.2 & 81.5 & 55.0 & 73.7 & 18.7 & 43.5 & 33.8 & 36.7 \\
			UDSX~\cite{UDSX}$_\textit{Neurocomp.'24}$ &  & 60.4 & 83.2 & 55.8 & 74.7 & 20.2 & 47.6 & 37.2 & 38.9 \\ \midrule
			RILL$\dagger$~\cite{RILL}$_\textit{TCSVT'23}$ & \multirow{5}{*}{Normalization} & 63.5 & 84.1 & 56.9 & 74.8 & 20.7 & 45.8 & 38.7 & 40.1 \\
			DCCL~\cite{DCCL}$_\textit{TCSVT'23}$ &  & 63.2 & 85.3 & 60.0 & 75.9 & 20.5 & 47.4 & 38.7 & 40.3 \\
			STL~\cite{STL}$_\textit{ICME'24}$ &  & 63.3 & 85.7 & 57.1 & 74.6 & 21.3 & 45.1 & 40.4 & 41.9 \\
			IGMG$^*$~\cite{IGMG}$_\textit{CVIU'24}$ &  & 57.1 & 83.1 & 57.7 & 76.2 & 19.7 & 41.6 & 36.3 & 37.3 \\
			SVIL~\cite{SVIL}$_\textit{TOMM'24}$ &  & 65.0 & 86.1 & \underline{61.7} & \underline{77.3} & 19.9 & 44.6 & \underline{43.2} & \underline{44.1} \\ \midrule
			CLIP-FGDI$^*$~\cite{CLIP-FGDI}$_\textit{TIFS'25}$ & \multirow{3}{*}{VLM-based} & \textcolor{gray}{46.3} & \textcolor{gray}{77.9} & \textcolor{gray}{51.0} & \textcolor{gray}{70.4} & \textcolor{gray}{28.8} & \textcolor{gray}{56.7} & \textcolor{gray}{34.4} & \textcolor{gray}{35.6} \\
			DCAC~\cite{DCAC}$_\textit{Sensors'25}$ &  & 56.7 & 80.0 & 58.9 & 75.4 & 27.5 & 56.7 & 42.5 & 43.6 \\
			MUVA$_\textit{Ours}$ &  & \underline{66.7} & 86.1 & \bf 63.8 & \bf 79.4 & \bf 34.5 & \bf 63.1 & \bf 48.6 & \bf 49.6 \\ \bottomrule
	\end{tabular}}
\end{table*}

\subsubsection{Multi-source DG Re-ID}
As illustrated in Table~\ref{tab:msdg_sota}, extensive comparisons are conducted under multi-source protocols. Overall, our approach achieves substantial and balanced improvements across all target domains. QAConv-MS~\cite{QAConv-MS}, which benefits from multi-scale image matching with a larger image input size, attains the best Rank-1 performance on target domain MA. However, it fails to sustain its superiority on the target domains MS and C3. In contrast, our approach consistently enhances generalization capability across all target domains without relying on additional training data or larger input sizes. Notably, it also outperforms ISR~\cite{ISR} on the most challenging target domain MS by 4.2\% in mAP and 6.7\% in Rank-1. Furthermore, when compared to the recently proposed SVIL~\cite{SVIL}, which achieves remarkable performance on MA, D and C3 with results similar to ours, our MUVA model surpasses it on dataset MS by a substantial margin of 14.6\% in mAP and 18.5\% in Rank-1. As for recent methods CLIP-FGDI~\cite{CLIP-FGDI} and DCAC~\cite{DCAC} which also employ vision-language models, our approach exceeds them consistently under all protocols. These results further underscore the effectiveness of our approach.

\begin{table}[t]
	\centering
	\caption{Comparison with state-of-the-art Re-ID methods on source domains. $*$ indicates that the input image size is larger than $256\times 128$. Note that results on CUHK03-NP~\cite{re-ranking} are not presented due to lack of experiments from recent works. Best and suboptimal results are emphasized in bold and underline, respectively.}
	\label{tab:supervised_sota}
	\resizebox{\linewidth}{!}{
		\begin{tabular}{lcccccc}
			\toprule
			\multirow{2}{*}{Model} & \multicolumn{2}{c}{Market1501} & \multicolumn{2}{c}{DukeMTMC-reID} & \multicolumn{2}{c}{MSMT17} \\ \cmidrule{2-7} 
			& mAP & Rank-1 & mAP & Rank-1 & mAP & Rank-1 \\ \midrule
			TransReID$_{256\times 128}$~\cite{TransReID}$_\textit{ICCV'21}$ & 88.9 & 95.2 & \underline{82.0} & \underline{90.7} & 67.4 & 85.3 \\
			DCAL~\cite{DCAL}$_\textit{CVPR'22}$ & 87.5 & 94.7 & 80.1 & 89.0 & 64.0 & 83.1 \\
			AAformer$^*$~\cite{AAformer}$_\textit{TNNLS'23}$ & 88.0 & 95.4 & 80.9 & 90.1 & 65.6 & 84.4 \\
			CLIP-ReID~\cite{CLIP-ReID}$_\textit{AAAI'23}$ & \underline{89.6} & \underline{95.5} & \textbf{82.5} & 90.0 & \underline{73.4} & 88.7 \\
			PromptSG~\cite{PromptSG}$_\textit{CVPR'24}$ & \bf 94.6 & \bf 97.0 & 81.6 & \bf 91.0 & \bf 87.2 & \bf 92.6 \\
			\midrule
			MUVA$_\textit{Ours}$ & 88.2 & 94.9 & 80.1 & 89.9 & 72.2 & \underline{89.0} \\ \bottomrule
	\end{tabular}}
\end{table}

\subsubsection{Performance on Source Domains}

In Table~\ref{tab:supervised_sota}, we validate the performance of MUVA on source domains by comparing it with five state-of-the-art Re-ID methods. Except PromptSG~\cite{PromptSG}, which extracts multi-modal features with both image and text inputs via a deeper model, the proposed MUVA demonstrates narrow performance gaps against other methods on all three datasets. This indicates that MUVA also performs well on source domains.

\subsection{Ablation Studies}
\subsubsection{Effectiveness of the Multiple Granularities}
\begin{table}[t]
	\centering
	\caption{Ablations on the effectiveness of different granularity combinations. ``G'', ``H'', ``U'' and ``L'' indicate global, head, upper body and leg regions, respectively. Best results are emphasized in bold.}
	\label{tab:ablation_granularity}
	\resizebox{\linewidth}{!}{
		\begin{tabular}{lcccccccc}
			\toprule
			\multirow{2}{*}{Model} & \multicolumn{4}{c}{Granularity} & \multicolumn{2}{c}{MA$\to$MS} & \multicolumn{2}{c}{MS$\to$MA} \\ \cmidrule{2-9} 
			& G & H & U & L & mAP & Rank-1 & mAP & Rank-1 \\
			\midrule
			MUVA (w/ G) & \checkmark &  &  &  & 26.0 & 54.7 & 51.9 & 77.4 \\
			MUVA (w/ G+H) & \checkmark & \checkmark &  &  & 25.0 & 53.7 & 53.4 & 78.6 \\
			MUVA (w/ G+U) & \checkmark &  & \checkmark &  & 26.4 & 55.8 & 52.9 & 78.8 \\
			MUVA (w/ G+L) & \checkmark &  &  & \checkmark & 27.3 & 56.2 & 56.7 & 81.6 \\
			MUVA (w/ G+H+U) & \checkmark & \checkmark & \checkmark &  & 26.9 & 56.1 & 55.5 & 79.9 \\
			MUVA (w/ G+H+L) & \checkmark & \checkmark &  & \checkmark & 27.0 & 56.0 & 56.8 & 81.1 \\
			MUVA (w/ G+U+L) & \checkmark &  & \checkmark & \checkmark & 27.2 & 56.0 & 57.1 & 81.1 \\
			MUVA (w/ G+H+U+L) & \checkmark & \checkmark & \checkmark & \checkmark & \bf 27.6 & \bf 56.8 & \bf 59.5 & \bf 82.7 \\
			\bottomrule
	\end{tabular}}
\end{table}
To validate the effectiveness of the multi-grained vision-language prompt learning, we conduct ablation studies on different combinations of part granularities using two single-source generalization protocols MA$\to$MS and MS$\to$MA. These protocols are also applied in subsequent ablation studies. The results are presented in Table~\ref{tab:ablation_granularity}. We denote the global, head, upper body and leg regions as ``G'', ``H'', ``U'' and ``L'' respectively. The experiments demonstrate that even coarse-grained vision-language alignment can produce a competitive baseline model ``MUVA (w/ G)'' with fair generalization performances of 26.0\% mAP and 54.7\% Rank-1 for MA$\to$MS, and 51.9\% mAP and 77.4\% Rank-1 for MS$\to$MA.

The generalization capability is gradually enhanced with the incorporation of more granularities. When parts are introduced separately (\eg, ``G+H'', ``G+U'' and ``G+L''), the refinements are limited. The best performance among these is achieved by introducing the leg region, with 27.3\% mAP and 56.2\% Rank-1 for MA$\to$MS, and 56.7\% mAP and 81.6\% Rank-1 for MS$\to$MA. This indicates that the leg region may provide more informative fine-grained semantics. Combinations of different parts (\eg, ``G+H+U'', ``G+H+L'' and ``G+U+L'') further improve generalization on MS$\to$MA, achieving 57.1\% mAP and 81.1\% Rank-1 for the best. Ultimately, the variant ``MUVA (w/ G+H+U+L)'' that includes all granularities, achieves the best performance of 27.6\% mAP and 56.8\% Rank-1 for MA→MS, and 59.5\% mAP and 82.7\% Rank-1 for MS→MA. This demonstrates that the synergy of multiple granularities significantly enhances generalization capability.

\begin{table}[t]
	\centering
	\caption{Ablations on the effectiveness of the language modality. ``vision-only'' indicates that the model is only trained with the image encoder. ``vision-language'' indicates that the model is trained by vision-language alignment. Best results are emphasized in bold.}
	\label{tab:ablation_language_modality}
	\resizebox{\linewidth}{!}{
		\begin{tabular}{lccccc}
			\toprule
			\multirow{2}{*}{Model} & \multirow{2}{*}{Granularity} & \multicolumn{2}{c}{MA$\to$MS} & \multicolumn{2}{c}{MS$\to$MA} \\ \cmidrule{3-6} 
			& & mAP & Rank-1 & mAP & Rank-1 \\
			\midrule
			MUVA (vision-only) & \multirow{2}{*}{Single-grained} & 25.2 & 53.8 & 52.6 & 77.1 \\
			MUVA (vision-language) & & 26.0 & 54.7 & 51.9 & 77.4 \\ \midrule
			MUVA (vision-only) & \multirow{2}{*}{Multi-grained} & 26.7 & 55.2 & 56.8 & 80.8 \\
			MUVA (vision-language) & & \bf 27.6 & \bf 56.8 & \bf 59.5 & \bf 82.7 \\
			\bottomrule
	\end{tabular}}
\end{table}

\subsubsection{Effectiveness of the Language Modality}
Table~\ref{tab:ablation_language_modality} presents an investigation into the effectiveness of incorporating the language modality. The model labeled with ``MUVA (vision-only)'' represents a simplified version, where stage-1 training and stage-2 alignment losses ($\mathcal{L}_{cmp}$ and $\mathcal{L}_{i2tce}$) are not included to introduce language knowledge. In contrast, ``MUVA (vision-language)'' denotes the complete model adopting vision-language alignment. Initially, the ablation is conducted under single-grained settings. On MA$\to$MS, the introduction of language modality appears slightly effective with 0.8\% mAP and 0.9\% Rank-1 improvements. However, on MS$\to$MA the Rank-1 is merely improved by 0.3\% and a drop of 0.7\% mAP is observed simultaneously. This reflects that the global-level alignment is not accurate enough. Some ID-discriminative nuances are ignored. After the use of multi-grained alignment, the improvement becomes more obvious with 2.7\% mAP and 1.9\% Rank-1 on MS$\to$MA. On more difficult MA$\to$MS protocol, the enhancement also raises to 0.9\% mAP and 1.6\% Rank-1. Apparently, the adoption of multi-grained features further refines the effectiveness of the vision-language alignment.

Moreover, when language information is not employed (compare ``MUVA (vision-only)'' models), we observe that the multi-grained model brings limited rises of 1.5\% mAP and 1.4\% Rank-1 on MA$\to$MS, and 4.2\% mAP and 3.7\% Rank-1 on MS$\to$MA. After adopting language information (compare ``MUVA (vision-language)'' models), the performance gains enlarge to 1.6\% mAP and 2.1\% Rank-1 on MA$\to$MS, and 7.6\% mAP and 5.3\% Rank-1 on MS$\to$MA, which illustrates that the multi-grained model tends to unleash greater potentials if it is assisted with language information, suggesting that it is the synergy of multiple granularity and language modality that leads to better performance.

\begin{table}[t]
	\centering
	\caption{Ablations on different prompt designs. ``PAP'' and ``DSP'' indicate the uses of part-agnostic prompt and domain-specific prompt, respectively. ``original'' indicates the use of original prompt in our model. Best results are emphasized in bold.}
	\label{tab:ablation_prompt_designs}
	\resizebox{\linewidth}{!}{
		\begin{tabular}{lcccc}
			\toprule
			\multirow{2}{*}{Model} & \multicolumn{2}{c}{MA$\to$MS} & \multicolumn{2}{c}{MS$\to$MA} \\ \cmidrule{2-5} 
			& mAP & Rank-1 & mAP & Rank-1 \\ \midrule
			MUVA (w/ PAP) & \bf 27.7 & 56.7 & 58.5 & 82.2 \\
			MUVA (w/ DSP) & 27.4 & 56.4 & 58.5 & 82.0 \\
			MUVA (original) & 27.6 & \bf 56.8 & \bf 59.5 & \bf 82.7 \\ \bottomrule
	\end{tabular}}
\end{table}

\subsubsection{Ablations on Prompt Designs}
In Table~\ref{tab:ablation_prompt_designs}, we study the effectiveness of different prompt designs. ``MUVA (original)'' indicates the use of original prompt as introduced previously in this paper. ``MUVA (w/ PAP)'' indicates the use of part-agnostic prompt, which removes all part names in the template to exclude the knowledge of body parts. Without textual part information, there are no salient and consistent influences found on MA$\to$MS except 0.1\% fluctuation on each metric. But a drop of 1.0\% mAP and 0.5\% Rank-1 is observed on MS$\to$MA, suggesting that specific words describing parts encourage better multi-grained alignment. ``MUVA (w/ DSP)'' indicates the use of domain-specific prompt, which adds more descriptions about the domain's style (\eg illumination, resolution, etc.) according to corresponding dataset. In this variant, we want to validate if prior domain information is helpful. However, we find it impairs the performances with a drop of 0.2\% mAP and 0.4\% Rank-1 on MA$\to$MS, and 1.0\% mAP and 0.7\% Rank-1 on MS$\to$MA. This indicates that the prompt should be domain-agnostic to avoid distracting the learning of generalizable ID features across domains.

\begin{table}[t]
	\centering
	\caption{Ablations on the effectiveness of AM-MSA module. ``AM-MSA'' indicates our approach using the AM-MSA module to explore parts. ``stripe-like'' indicates the use of traditional stripe-like partition. Best results are emphasized in bold.}
	\label{tab:ablation_partition}
	\resizebox{\linewidth}{!}{
		\begin{tabular}{lcccc}
			\toprule
			\multirow{2}{*}{Model} & \multicolumn{2}{c}{MA$\to$MS} & \multicolumn{2}{c}{MS$\to$MA} \\ \cmidrule{2-5} 
			& mAP & Rank-1 & mAP & Rank-1 \\
			\midrule
			MUVA (stripe-like) & 25.8 & 54.3 & 55.5 & 80.1 \\
			MUVA (AM-MSA) & \bf 27.6 & \bf 56.8 & \bf 59.5 & \bf 82.7 \\
			\bottomrule
	\end{tabular}}
\end{table}

\subsubsection{Effectiveness of the AM-MSA Module}
In Table~\ref{tab:ablation_partition}, we evaluate the effectiveness of the proposed AM-MSA module. We compare our approach, denoted as ``MUVA (AM-MSA)'', with the traditional stripe-like partition strategy, denoted as ``MUVA (stripe-like)''. In our approach, multi-grained features are extracted using attention masks predicted by the AM-MSA module. Conversely, in the traditional approach, stripe-like equally-partitioned attention masks are utilized. Both strategies employ the same granularities for head, upper body and legs. The results indicate that our proposed method achieves superior generalization performance compared to the stripe-like partition strategy, significantly demonstrating its effectiveness.

\begin{table}[t]
	\centering
	\caption{Ablations on pseudo label mask calibration. Best results are emphasized in bold.}
	\label{tab:ablation_calibration}
	\resizebox{\linewidth}{!}{
		\begin{tabular}{lcccc}
			\toprule
			\multirow{2}{*}{Model} & \multicolumn{2}{c}{MA$\to$MS} & \multicolumn{2}{c}{MS$\to$MA} \\ \cmidrule{2-5} 
			& mAP & Rank-1 & mAP & Rank-1 \\
			\midrule
			MUVA (w/o calibration) & 27.3 & 56.1 & 57.7 & 81.7 \\
			MUVA (w/ calibration) & \bf 27.6 & \bf 56.8 & \bf 59.5 & \bf 82.7 \\
			\bottomrule
	\end{tabular}}
\end{table}

Furthermore, Table~\ref{tab:ablation_calibration} examines the effectiveness of the calibration scheme used to mitigate the hallucination issues associated with the visual grounding expert. The model denoted as ``MUVA (w/o calibration)'' is trained using the raw pseudo label masks predicted by the expert. This model exhibits suboptimal performance due to inaccurate part location. In contrast, the model ``MUVA (w/ calibration)'' utilizes calibrated pseudo label masks, resulting in improved performance. This comparison underscores the effectiveness and necessity of the calibration process in enhancing the final performance of the model.

\subsubsection{Ablations on Structure of the RMP Module}
Table~\ref{tab:ablation_rmp_structure} explores the impact of different RMP module structures on the generalization performance. Drawing inspirations from the successful practices in residual networks~\cite{ResNet} and transformer architectures~\cite{ViT,Transformer}, we experiment with incorporating residual connections and layer normalization around the cross-attention within the RMP module. Various structural configurations are evaluated to determine their contributions to generalization enhancements. Specifically, ``LN'' denotes the application of layer normalization, while ``RC'' signifies the use of a residual connection from inputs to outputs of the cross-attention.

\begin{table}[t]
	\centering
	\caption{Ablations on the structure of the RMP module. ``LN'' indicates the layer normalization. ``RC'' indicates the use of residual connection from inputs to outputs of the cross-attention. Best results are emphasized in bold.}
	\label{tab:ablation_rmp_structure}
	\resizebox{\linewidth}{!}{
		\begin{tabular}{lcccccc}
			\toprule
			\multirow{2}{*}{Model} & \multicolumn{2}{c}{RMP Module} & \multicolumn{2}{c}{MA$\to$MS} & \multicolumn{2}{c}{MS$\to$MA} \\ \cmidrule{2-7} 
			& LN & RC & mAP & Rank-1 & mAP & Rank-1 \\
			\midrule
			MUVA (RMP w/o LN+RC) &  &  & 26.9 & 55.6 & 58.4 & 82.2 \\
			MUVA (RMP w/ LN) & \checkmark &  & 27.4 & 56.1 & 58.7 & \bf 82.8 \\
			MUVA (RMP w/ RC) &  & \checkmark & 27.4 & 56.6 & 59.2 & 82.6 \\
			MUVA (RMP w/ LN+RC) & \checkmark & \checkmark & \bf 27.6 & \bf 56.8 & \bf 59.5 & 82.7 \\
			\bottomrule
	\end{tabular}}
\end{table}

The results indicate that both layer normalization and residual connections individually contribute to enhancing the generalization capability. Specifically, residual connections appear to provide a more substantial improvement in generalization. However, the combined use of layer normalization and residual connections yields superior results compared to employing either technique alone, with the exception of a minor decline in Rank-1 performance for the MS$\to$MA protocol. This suggests that the model benefits the most from the synergistic interaction between these two designs.

\subsubsection{Ablations on Random Cropping and Random Erasing Augmentations}

\begin{table}[t]
	\centering
	\caption{Ablations on the effectiveness of adopting random cropping and random erasing~\cite{random_erasing} as augmentations. ``RCRE'' indicates two augmentations. Best results are emphasized in bold.}
	\label{tab:ablation_rcre}
	\resizebox{\linewidth}{!}{
		\begin{tabular}{lcccc}
			\toprule
			\multirow{2}{*}{Model} & \multicolumn{2}{c}{MA$\to$MS} & \multicolumn{2}{c}{MS$\to$MA} \\ \cmidrule{2-5} 
			& mAP & Rank-1 & mAP & Rank-1 \\ \midrule
			MUVA (w/ RCRE) & 25.9 & 55.1 & 57.4 & 81.7 \\
			MUVA (w/o RCRE) & \bf 27.6 & \bf 56.8 & \bf 59.5 & \bf 82.7 \\ \bottomrule
	\end{tabular}}
\end{table}

\begin{table}[t]
	\centering
	\caption{Ablations on the number of the cross-attention heads in the RMP module. Best results are emphasized in bold.}
	\label{tab:ablation_rmp_head}
	\resizebox{\linewidth}{!}{
		\begin{tabular}{ccccc}
			\toprule
			\multirow{2}{*}{Number of Heads} & \multicolumn{2}{c}{MA$\to$MS} & \multicolumn{2}{c}{MS$\to$MA} \\ \cmidrule{2-5} 
			& mAP & Rank-1 & mAP & Rank-1 \\
			\midrule
			1 & 27.4 & 56.1 & 58.2 & 81.7 \\
			2 & \bf 27.6 & \bf 56.8 & 58.5 & 82.2 \\
			4 & 27.4 & 55.9 & 58.7 & 82.4 \\
			8 & 27.2 & 56.0 & \bf 59.5 & \bf 82.7 \\
			16 & 26.9 & 55.7 & 59.2 & 82.7 \\
			\bottomrule
	\end{tabular}}
\end{table}

\begin{table*}[t]
	\centering
	\caption{Comparisons of different models on parameter scale, time latency per batch, GPU memory consumption and Re-ID performance. ``Train.'' and ``Infer.'' indicate training and inference, respectively. ``S1'' and ``S2'' indicate stage-1 and stage-2 training, respectively. The batch size is set to 1 for the VGE and 64 for others.}
	\label{tab:computational_cost}
	\resizebox{\textwidth}{!}{
		\begin{tabular}{lcccccccccccc}
			\toprule
			\multirow{2}{*}{Model} & \multicolumn{2}{c}{Parameters (M)} & \multicolumn{3}{c}{Latency (ms)} & \multicolumn{3}{c}{Memory (GB)} & \multicolumn{2}{c}{MA$\to$MS} & \multicolumn{2}{c}{MS$\to$MA} \\ \cmidrule{2-13} 
			& Train. & Infer. & S1 & S2 & Infer. & S1 & S2 & Infer. & mAP & Rank-1 & mAP & Rank-1 \\ \midrule
			PAT~\cite{PAT} & 87.1 & 86.5 & - & 364.3 & 23.2 & - & 5.7 & 1.1 & 18.2 & 42.8 & 45.8 & 71.6 \\
			CLIP-FGDI~\cite{CLIP-FGDI} & 129.5 & 86.2 & 100.9 & 273.1 & 22.1 & 1.6 & 6.5 & 0.8 & 20.2 & 47.0 & 44.7 & 73.0 \\
			\midrule
			MUVA (w/o AM-MSA) & 126.2 & 86.1 & 58.9 & 152.1 & 22.1 & 2.1 & 4.6 & 0.7 & 26.0 & 54.7 & 51.9 & 77.4 \\
			MUVA (w/o lang.) & 117.4 & 115.9 & - & 243.4 & 71.1 & - & 6.8 & 0.9 & 26.7 & 55.2 & 56.8 & 80.8 \\
			MUVA & 161.7 & 115.9 & 161.2 & 247.2 & 69.4 & 7.2 & 6.9 & 0.9 & 27.6 & 56.8 & 59.5 & 82.7 \\
			\midrule
			VGE (part locating only) & - & 16822.5 & - & - & 7243.6 & - & - & 34.3 & - & - & - & - \\
			\bottomrule
	\end{tabular}}
\end{table*}

We investigate the effectiveness of adopting random cropping and random erasing~\cite{random_erasing}, which have been utilized as default augmentations by many previous works. The goal of such augmentations is to diversify training samples, encouraging more robust feature learning. However, the generalization performance degenerates when these augmentations are adopted in our model, as presented in the Table~\ref{tab:ablation_rcre}. The reason is that cropping and erasing tend to randomly destroy body parts in the image, leading to the elimination of some local semantics. But our model is designed to perceive such fine-grained information. With incomplete local regions, the model fails to effectively align nuanced semantics between vision and language modalities, which finally impairs generalization performance. Therefore, we decide to disable these augmentations in our full model.

\subsubsection{Analysis of the Cross-Attention Heads in the RMP Module}

Table~\ref{tab:ablation_rmp_head} presents an analysis of the optimal number of cross-attention heads in the RMP module. Multiple attention heads are typically utilized to augment the representation capability of the attention mechanism~\cite{Transformer}. Our experimental findings indicate that the ideal number of attention heads correlates with the scale of the source domain training data. We use MA and MS to represent source domains of different scales, where MA represents smaller domains with fewer samples and IDs, while MS signifies larger domains with a greater number of samples and IDs. The results demonstrate that fewer attention heads are more appropriate for smaller domains, whereas larger domains benefit from a higher number of heads.

\begin{figure*}[t]
	\centering
	\includegraphics[width=\textwidth]{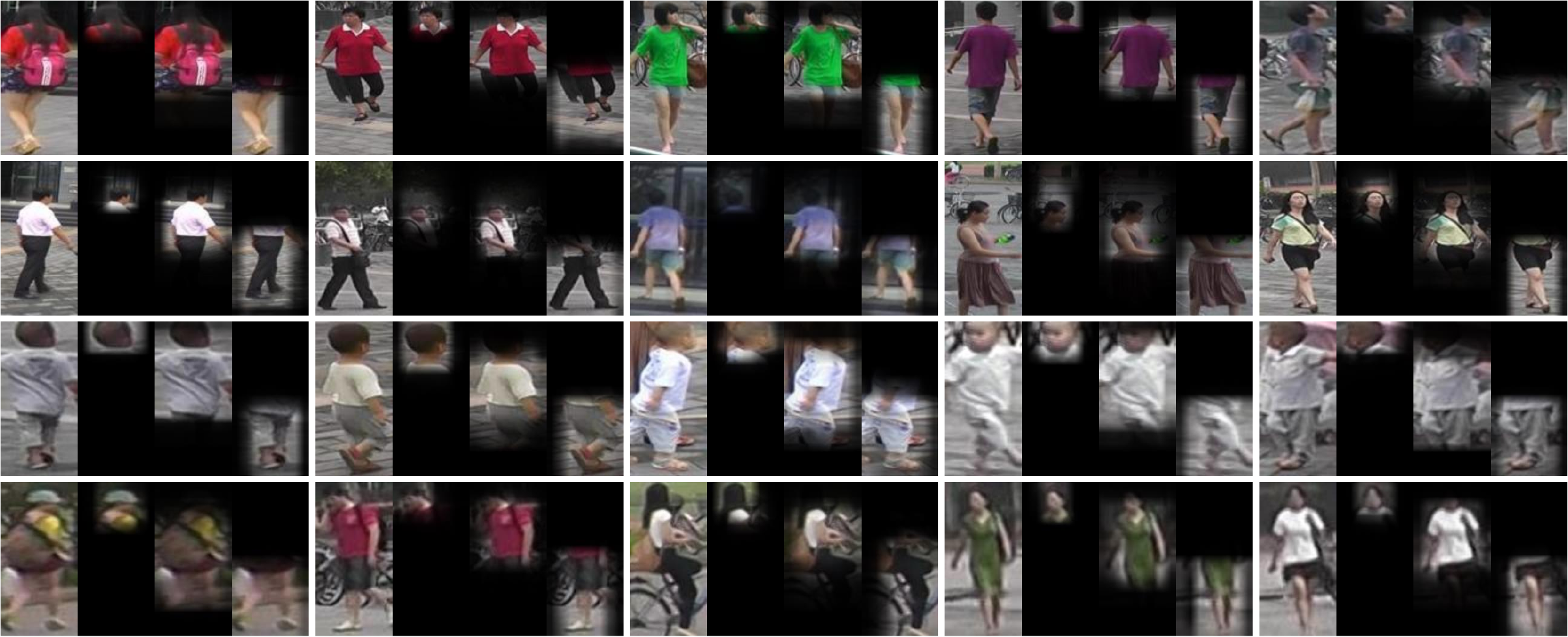}
	\caption{Visualization results of the part masks generated in the AM-MSA module. The model is trained on MSMT17~\cite{MSMT17} dataset and visualized on the query and gallery sets of Market1501~\cite{Market1501} dataset. For each sample, the original image and the masks for head, upper body and legs are presented from left to right. Brighter regions indicate higher model focuses.}
	\label{fig:vis_part_locating}
\end{figure*}

\begin{figure*}[t]
	\centering
	\includegraphics[width=\textwidth]{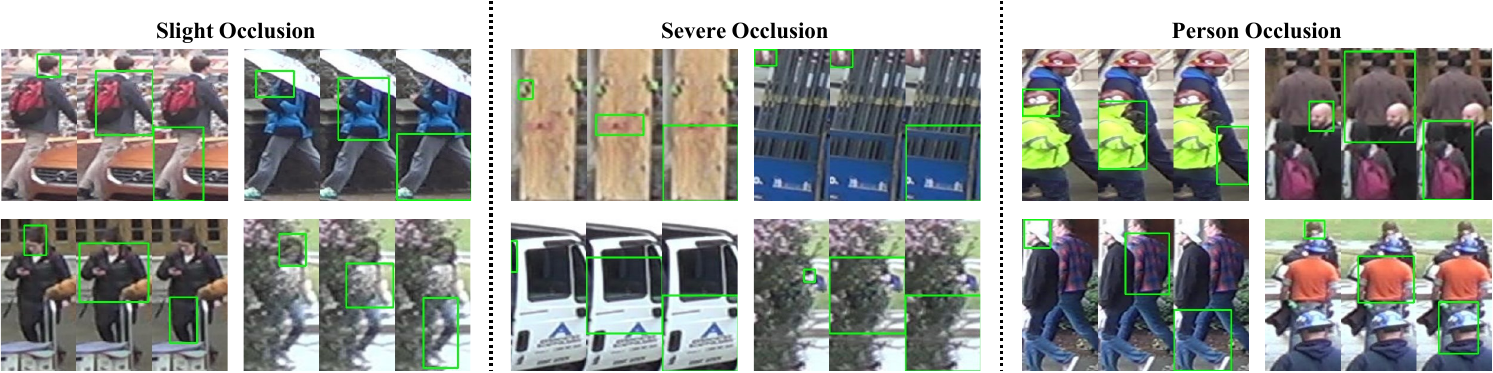}
	\caption{Some cases of visual grounding on occluded samples from DukeMTMC-reID~\cite{DukeMTMC-reID} dataset. Left group shows slight occlusion. Middle group shows severe occlusion. Right group shows occlusion caused by other people. In each sample, head, upper body and legs are detected from left to right, respectively.}
	\label{fig:vis_occlusion}
\end{figure*}

\subsection{Computational Cost}
In Table~\ref{tab:computational_cost}, we investigate the computational cost. To make a comprehensively comparison, we count the model parameter scale, time latency per batch, GPU memory consumption and Re-ID performance. CLIP-FGDI~\cite{CLIP-FGDI} is a similar model to ours, which also employs vision-language alignment but contains additional domain-invariant/relevant prompt learning. PAT~\cite{PAT} is a pure visual model, which learns generalizable features from part-level similarities across diverse IDs. Differently, our MUVA adopts both language and part designs. As a result, it takes more computational resources. But it brings remarkable DG Re-ID performance improvement with acceptable inference latency of 69.4ms per batch and 0.9GB GPU memory consumption in inference. This is because only the image encoder is required in inference, which significantly reduces the number of parameters.

We also compare the full model with two variants. ``MUVA (w/o AM-MSA)'' indicates that we only use single-grained features for vision-language alignment. ``MUVA (w/o lang.)'' indicates that we take multi-grained features but disable the use of language modality. Although the use of the AM-MSA module introduces more computational burden, it contributes the most to DG Re-ID performance. The use of language modality brings no obvious influences on time latency and memory consumption during inference but improves Re-ID results effectively. Ultimately, we compare our MUVA with the VGE on inference cost. The batch size of the VGE is set to 1 due to device restriction. But it consumes far more resources merely on part locating. It even oversteps the cost of entire feature extraction process of MUVA with a batch size of 64. Evidently, it is unsuitable to deploy the VGE in online Re-ID systems to extract fine-grained information. By contrast, our MUVA achieves a nice trade-off between performance and cost.

\subsection{Visualization Results}
To further substantiate the part locating capability of the AM-MSA module, we visualize the predicted masks on target domain in Figure~\ref{fig:vis_part_locating}. The intensity of the prediction scores after sigmoid activation is depicted by image brightness. Brighter regions indicate higher model focuses. The results reveal that the AM-MSA module effectively generalizes to unseen target domains and is resilient to different body poses. For smaller regions like head, our model achieves precise locating, thereby facilitating superior aggregation of fine-grained ID-related information. Additionally, we observe that the parts can be detected accurately even if the person is not compactly bounded in image, meaning that our model learns correct knowledge of body parts. We also find our model is robust to body part ratio and blur, as illustrated by children images and the images captured from far distance in the third and fourth rows, respectively.

\section{Limitations}

Although the proposed AM-MSA module and the VGE supervision endow our model with stronger capability of perceiving nuanced semantics, their robustness against occluded scenarios remains unknown. Thus, we investigate some occlusion cases during VGE-based pseudo label generation in Figure~\ref{fig:vis_occlusion}. When slight occlusion occurs, we find that the VGE still performs well to locate uncovered regions. Body parts are correctly detected even if they are partially occluded. This indicates that the VGE is robust to slight occlusion disturbance due to its remarkable capability obtained from pre-training. However, if a person is severely covered by obstacles, where the entire body is almost invisible, the VGE fails to recognize correct parts due to insufficient related pixels in images. In another case where a person is occluded by other people, the VGE is confused to determine the main person even though all body parts are detected. When this happens, body parts from different people may be incorrectly grouped, impairing the consistency of visual grounding for the same ID.

In future works, we plan to mitigate such problems by further unleashing the multi-modal understanding capability of the VGE. For example, the VGE can double-check its predictions with the images containing bounding boxes. Using proper prompts, the VGE can be asked to judge the type of regions inside the boxes. If some regions are considered as obstacles, they are discarded before loss computation. Similarly, the VGE can also be prompted to ignore the samples containing more than one person, preventing the inconsistent body part visual grounding.

\section{Conclusion}

In this work, we introduce a multi-grained vision-language alignment framework to improve DG Re-ID. Compared with single-grained alignment, our approach enhances the perception of nuanced ID differences by adopting learnable multi-grained prompts in language modality. To obtain fine-grained visual features, we propose an adaptively masked multi-head self-attention module supervised by an MLLM-based visual grounding expert to locate body parts. Through extensive experiments and ablation studies, our approach has demonstrated its effectiveness in DG Re-ID, achieving state-of-the-art performance. These results highlight the significant potential of adapting vision-language models for DG Re-ID and underscore the importance of incorporating multi-grained features to enhance generalization capability.

Broadly speaking, our work can be further extended to enhance related directions. Other object Re-ID tasks, such as animal or vehicle Re-ID, may transfer our model to more complicated domains by designing object-specific prompts. By adopting the VLM for video understanding and considering temporal information in prompts, our work can also be used to train generalizable video-based Re-ID models, which focus not only on spatio nuances but also on temporal dependencies.

{
    \small
    \bibliographystyle{ieeenat_fullname}
    \bibliography{main}
}


\end{document}